\RequirePackage{fix-cm}
\documentclass[twocolumn]{svjour3}    
\pdfoutput=1

\usepackage{graphicx}
\usepackage{natbib}

\usepackage{url}            
\usepackage{booktabs}       
\usepackage{amsfonts}       
\usepackage{nicefrac}       
\usepackage{microtype}      
\usepackage{graphicx}
\usepackage{enumitem}
\usepackage{bm}
\usepackage{nccmath}
\usepackage{boldline}
\usepackage{booktabs} 

\usepackage{algorithm}
\usepackage[noend]{algcompatible}

\usepackage{capt-of}
\usepackage{color,soul}
\usepackage[pagebackref=false,breaklinks=true,colorlinks,bookmarks=false,urlcolor=blue, citecolor=blue]{hyperref}

\usepackage{times}
\usepackage{epsfig}
\usepackage{multirow}
\usepackage{xcolor}
\usepackage{amsmath}
\usepackage{amssymb}
\usepackage{wrapfig}
\usepackage{comment}

\usepackage{textcomp}
\usepackage{bm}
\usepackage{soul}

\usepackage{makecell}
\usepackage{subfigure}
\usepackage{arydshln}
\usepackage{ulem}

\newcommand{\nicknameData}{SensatUrban}

\newcommand{\ie}{\textit{i}.\textit{e}., }
\newcommand{\eg}{\textit{e}.\textit{g}., }

\useunder{\underline}{\ul}{}

\usepackage{tikz}

\begin{document}

\title{SensatUrban: Learning Semantics from Urban-Scale Photogrammetric Point Clouds}

\author{Qingyong Hu \and Bo Yang \and Sheikh Khalid \and  Wen Xiao \and Niki Trigoni \and  Andrew Markham }

\institute{Qingyong Hu \and Niki Trigoni \and Andrew Markham \at
         Department of Computer Science,\\ 
         University of Oxford \\ 
         \email{firstname.lastname@cs.ox.ac.uk}
         \and
          Bo Yang \at
          Department of Computing, \\
          The Hong Kong Polytechnic University, \\
          \email{bo.yang@polyu.edu.hk}
         \and
         Sheikh Khalid \at
         Sensat Ltd. 
         \and
         Wen Xiao \at
         School of Engineering, \\
         Newcastle University \\
         \email{wen.xiao@newcastle.ac.uk}
}
\date{Received: date / Accepted: date}

\maketitle

\begin{abstract}
With the recent availability and affordability of commercial depth sensors and 3D scanners, an increasing number of 3D (\textit{i.e.}, RGBD, point cloud) datasets have been publicized to facilitate  research in 3D computer vision. However, existing datasets either cover relatively small areas or have limited semantic annotations. Fine-grained understanding of urban-scale 3D scenes is still in its infancy. In this paper, we introduce SensatUrban, an urban-scale UAV photogrammetry point cloud dataset consisting of nearly three billion points collected from three UK cities, covering 7.6 $km^2$. Each point in the dataset has been labelled with fine-grained semantic annotations, resulting in a dataset that is three times the size of the previous existing largest photogrammetric point cloud dataset. In addition to the more commonly encountered categories such as road and vegetation, urban-level categories including rail, bridge, and river are also included in our dataset. Based on this dataset, we further build a benchmark to evaluate the performance of state-of-the-art segmentation algorithms. In particular, we provide a comprehensive analysis and identify several key challenges limiting urban-scale point cloud understanding. The dataset is available at \url{http://point-cloud-analysis.cs.ox.ac.uk/}.

\end{abstract}

\section{Introduction}
\label{sec:intro}
Giving machines the ability to semantically interpret 3D scenes  is highly important for accurate 3D perception and scene understanding. This is also the prerequisite for numerous real-world applications such as object-level robotic grasping \cite{rao2010grasping}, scene-level robot navigation \cite{valada2017adapnet} and autonomous driving \cite{geiger2013vision}, or even large-scale urban 3D modeling, where autonomous machines are required to interact competently within our physical world. Although increasing research attention has been applied to this field, it remains challenging due to the high geometrical complexity of urban scenes, and limited high quality labelled data resources.

\begin{figure}[t]
\centering
\includegraphics[width=0.5\textwidth]{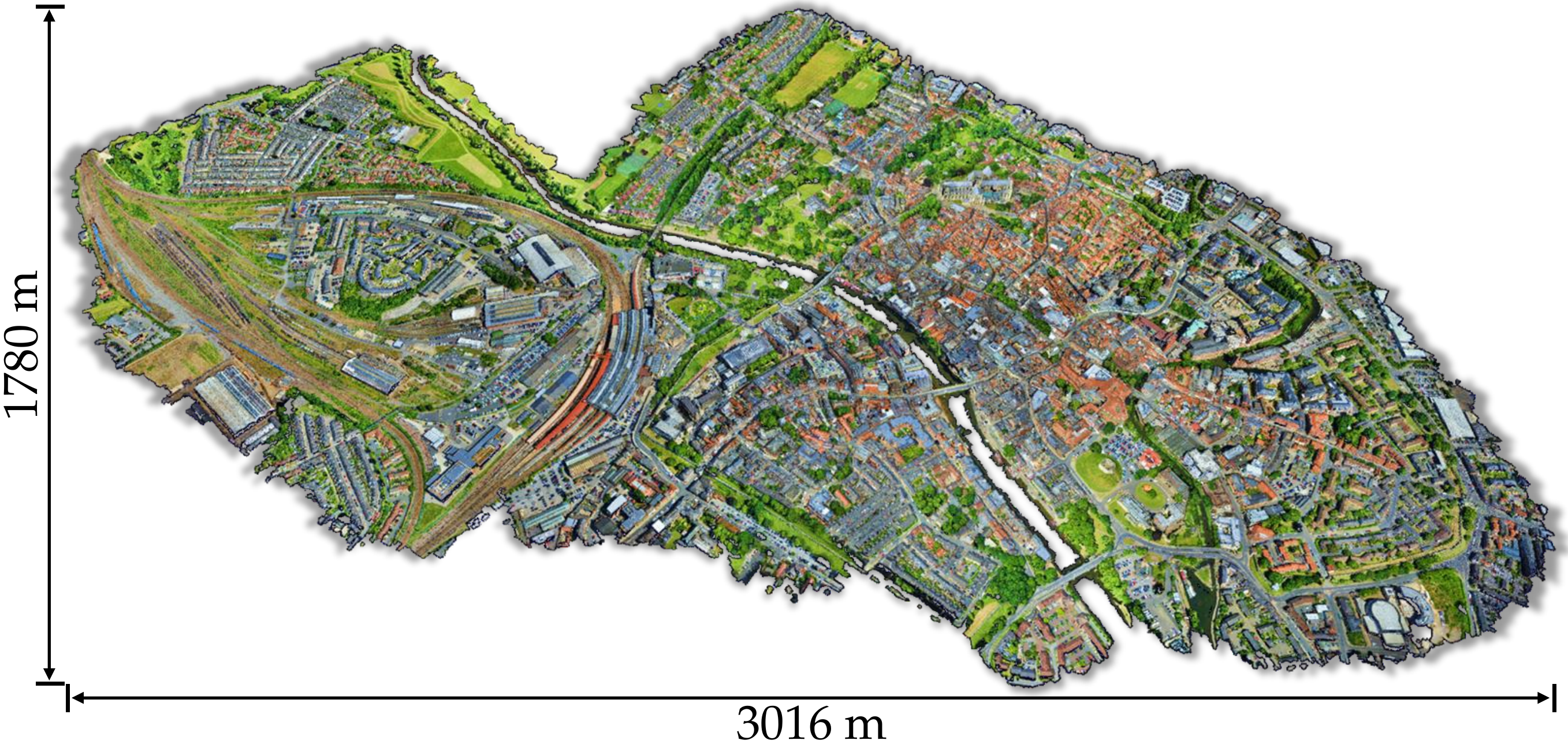}
\caption{This shows an example of urban-scale point clouds in our \nicknameData{} dataset. It is acquired from the city center of York through UAV photogrammetry. It has a spatial coverage of more than 3 square kilometers and represents a typical urban suburb.}\label{fig:urban-scale} 
\end{figure}

Recently, an increasingly number of sophisticated neural pipelines have been proposed based on different representations of 3D scenes, including: 1) 3D voxel-based methods such as SegCloud \cite{tchapmi2017segcloud}, SparseConvNet \cite{sparse}, MinkowskiNet \cite{4dMinkpwski}, PVCNN \cite{Point_voxel_cnn}, Cylinder3D \cite{cylinder3d} and 2) 2D projection-based approaches such as RangeNet++ \cite{rangenet++}, SalsaNext \cite{salsanext} and SqueezeSeg \cite{wu2018squeezeseg}, PolarNet \cite{zhang2020polarnet} and 3) recent point-based architectures \textit{e.g.} PointNet/PointNet++ \cite{qi2017pointnet,qi2017pointnet++}, PointCNN \cite{li2018pointcnn}, DGCNN \cite{dgcnn},  KPConv \cite{thomas2019kpconv}, RandLA-Net \cite{hu2019randla} and PointTransformer \cite{Point_transformer}.

The core of these techniques, however, relies heavily on the wide availability of large-scale and high-quality open datasets. The datasets provide realistic and diverse data resources and act as benchmarks to fairly evaluate and compare the performance of different algorithms. Existing representative 3D data repositories can be generally classified as: 1) object-level 3D models such as ModelNet \cite{3d_shapenets}, ShapeNet \cite{chang2015shapenet} and ScanObjectNN \cite{scanobjectnn}, 2) indoor scene-level 3D scans, \eg S3DIS \cite{2D-3D-S}, ScanNet \cite{scannet}, Matterport3D \cite{Matterport3D} and SceneNN \cite{scenenn}, and 3) outdoor roadway-level 3D point clouds including Semantic3D \cite{Semantic3D}, SemanticKITTI \cite{behley2019semantickitti}, NPM3D \cite{NPM3D}, and Toronto3D \cite{Toronto3D}. 

However, there is no large-scale photorealistic 3D point cloud dataset available for fine-grained semantic understanding of urban scenarios. Moreover, it remains an open question as to whether existing techniques can be scaled to these urban-scale point clouds. \textbf{Firstly}, in contrast to existing datasets for objects, rooms, or streets which are usually less than 200$m$ in scale, the urban-scale datasets collected by aerial platforms typically span extremely wide areas, e.g. kilometres. How to efficiently and effectively preprocess massive point sets (\eg over $10^8$) to feed into neural networks is a particular question of interest. \textbf{Secondly}, existing photogrammetric mapping techniques allow reconstructing photorealistic colorized point clouds. Along with the 3D spatial coordinates, is the inclusion of appearance beneficial to semantic understanding and what is the impact if any? \textbf{Thirdly}, real-world urban scenarios usually exhibit extreme class imbalance. The majority of points are dominated by categories such as ground and vegetation, while the critical categories such as rail and water only occupy a small proportion of the total number of points. \textbf{Fourthly}, and potentially most importantly, what is the generalization performance of existing deep neural networks? Can a trained model be well-generalized to unseen data, particularly from a different region? or even generalized to different dataset? \textbf{Lastly}, is it possible to learn semantics with sparser labels? How can we unleash the potential of self-supervised pre-training and semi-supervised learning on 3D point clouds?

In this paper, we take a step towards resolving the above issues. In particular, we first build a UAV photogrammetric point cloud dataset called \textbf{\nicknameData{}} for urban-scale 3D semantic understanding. This dataset covers 7.6 $km^2$ of urban areas in three UK cities \textit{i.e.}, Birmingham, Cambridge, and York (Figure \ref{fig:urban-scale}), along with nearly 3 billion richly annotated 3D points. Each point in the Birmingham and Cambridge set is enriched with one of 13  predefined semantic categories such as \textit{ground}, \textit{vegetation}, \textit{car}, \textit{etc.}, while the points in York remain unlabeled for potential semi-supervised researches. The 3D point clouds are reconstructed from highly overlapped sequential aerial images captured by a professional-grade UAV mapping system. For more detailed data acquisition pipelines, please refer to Section \ref{sec:dataset}. Compared with existing 3D datasets, the uniqueness of our \nicknameData{} lies in two aspects:

\begin{itemize}[leftmargin=*]
\setlength{\itemsep}{0pt}
\setlength{\parsep}{0pt}
\setlength{\parskip}{0pt}
\item \textbf{Urban-scale spatial coverage}. In contrast to  existing datasets which mainly focus on objects \cite{3d_shapenets,chang2015shapenet}, rooms \cite{scenenn,2D-3D-S,scannet} and roadways \cite{Semantic3D,behley2019semantickitti, NPM3D, Toronto3D}, the point clouds in our \nicknameData{} dataset continuously cover several square kilometers of real-world urban areas, opening up new opportunities towards urban-scale applications such as smart cities, and national infrastructure planning and management.
\item \textbf{Photorealistic and dense point clouds}. Our dataset is reconstructed from high-resolution aerial images captured by professional calibrated cameras.  Unique aerial images from nadir (top-down) and oblique perspectives for the entire landscape of cities are also provided for optimized and high-quality point clouds. Naturally, the geometric patterns, textures, natural colors, point density, and distributions are distinct from existing LiDAR-based datasets. 
\end{itemize}
\vspace{-0.1cm}

Based on the proposed \nicknameData{} dataset, we further highlight several new challenges faced by generalizing existing segmentation algorithms to urban-scale point clouds in Section \ref{sec:challenges}. In particular, these challenges include urban-scale data preparation, the usage of color information, learning from extremely imbalanced class distribution, cross-city generalization, and weakly and self-supervised learning from urban-scale point clouds. Note that, this paper does not aim to fully tackle these challenges, but to unveil them and provide insights to the community for future exploration.

To summarize, the main contributions of this paper are as follows:
\begin{itemize}[leftmargin=*]
\setlength{\itemsep}{0pt}
\setlength{\parsep}{0pt}
\setlength{\parskip}{0pt}
\item We propose a new urban-scale photogrammetric point cloud dataset for 3D semantic understanding, with an unprecedented spatial coverage at fine scale and rich semantic annotations.
\item We provide a comprehensive benchmark for semantic segmentation of urban-scale point clouds. Extensive experimental results of different state-of-the-art approaches are provided, with detailed discussions and analysis.
\item We highlight several unique challenges faced by generalizing existing neural pipelines to extremely large-scale point clouds, and provide an in-depth outlook of the future directions of 3D semantic learning.
\end{itemize}

A preliminary version of this work has been published in \cite{hu2020towards}, this journal extension particularly provides more details with regards to the data collection and point cloud reconstruction, additional experimental results and analysis in cross-dataset generalization, and weakly supervised semantic segmentation of urban-scale point clouds. In addition, the first challenge on large-scale point cloud analysis for urban scene understanding held in ICCV 2021 is based on this dataset. For more details, please refer to \url{https://urban3dchallenge.github.io/}.

\begin{table*}[t]
\centering
\caption{Comparison with existing representative 3D point cloud datasets. \protect\textsuperscript{1}The spatial size (Area/Length) in the dataset, m: meter. Note that, we use distance, instead of area size for outdoor roadway-level datasets. \protect\textsuperscript{2}The number of classes used for evaluation and the number of sub-classes annotated in brackets. MLS: Mobile Laser Scanning system, TLS: Terrestrial Laser Scanning system, ALS: Aerial Laser Scanning system. Note that, our dataset has total spatial coverage of 7.6 square kilometers, with nearly 3 billion richly annotated points with 13 semantic categories.}\label{tab:Overview-dataset-1}
\resizebox{0.95\textwidth}{!}{%
\begin{tabular}{crcrcccc}
\Xhline{2.0\arrayrulewidth} 
 & \#Name and Reference & \#Year & \#Spatial size\textsuperscript{1} & \#Classes\textsuperscript{2} & \#Points & \#RGB & \#Sensors\tabularnewline
\Xhline{1.25\arrayrulewidth} 
\multirow{4}{*}{Object-level} & ModelNet \cite{3d_shapenets} & 2015 & - & 40 & - & No & Synthetic \tabularnewline
& ShapeNet \cite{chang2015shapenet} & 2015 & - & 55 & - & No & Synthetic \tabularnewline
 & PartNet \cite{mo2019partnet} & 2019 & - & 24 & - & No & Synthetic\tabularnewline
 & ScanObjectNN \cite{mo2019partnet} & 2019 & - & 15 & - & Yes & RGB-D\tabularnewline
\Xhline{1.25\arrayrulewidth} 
\multirow{2}{*}{%
\begin{tabular}{c}
Indoor\tabularnewline
Scene-level\tabularnewline
\end{tabular}} & S3DIS \cite{2D-3D-S} & 2017 & 6$ \times 10^{3}m^2$ & 13 (13) & 273M & Yes & Matterport\tabularnewline
 & ScanNet \cite{scannet} & 2017 & 1.13$ \times 10^{5}m^2$ & 20 (20) & 242M & Yes & RGB-D\tabularnewline
\Xhline{1.25\arrayrulewidth} 

\multirow{6}{*}{%
\begin{tabular}{c}
Outdoor\tabularnewline
Roadway-level\tabularnewline
\end{tabular}} & Paris-rue-Madame \cite{paris-rue-madame} & 2014 & 0.16$\times 10^{3} \ m$ & 17 & 20M & No & MLS\tabularnewline
 & IQmulus \cite{IQmulus} & 2015 & 10$\times 10^{3} \ m$ & 8 (22) & 300M & No & MLS\tabularnewline
 & Semantic3D \cite{Semantic3D} & 2017 & - & 8 (9) & 4000M & Yes & TLS\tabularnewline
 & Paris-Lille-3D \cite{NPM3D} & 2018 & 1.94$\times 10^{3} \ m$ & 9 (50) & 143M & No & MLS\tabularnewline
 & SemanticKITTI \cite{behley2019semantickitti} & 2019 & 39.2$\times 10^{3} \ m$ & 25 (28) & 4549M & No & MLS\tabularnewline
 & Toronto-3D \cite{Toronto3D} & 2020 & 1$\times 10^{3} \ m$ & 8 (9) & 78.3M & Yes & MLS\tabularnewline
\Xhline{1.25\arrayrulewidth} 

\multirow{6}{*}{Urban-level} 
& ISPRS \cite{rottensteiner2012isprs} & 2012 & - & 9 & 1.2M & No & ALS\tabularnewline
& DublinCity \cite{zolanvari2019dublincity} & 2019 & 2$ \times 10^{6}m^2$ & 13 & 260M & No & ALS\tabularnewline
& DALES \cite{varney2020dales} & 2020 & 10 $ \times 10^{6}m^2$  & 8 (9) & 505M & No & ALS\tabularnewline
& LASDU \cite{ye2020lasdu} & 2020 & 1.02 $ \times 10^{6}m^2$  & 5 & 3.12M & No & ALS\tabularnewline
& Campus3D \cite{li2020campus3d} & 2020 & 1.58 $ \times 10^{6}m^2$ & 24 & 937.1M & Yes & UAV Photogrammetry\tabularnewline
 & \textbf{\textcolor{black}{\nicknameData{} (Ours)}} & 2020 & 7.64 $ \times 10^{6}m^2$ & 13 (31) & 2847M & Yes & UAV Photogrammetry\tabularnewline
\Xhline{2.0\arrayrulewidth} 
\end{tabular}}

\end{table*}

\section{Related Work}
\label{sec:liter}

\subsection{Datasets for 3D Scene Understanding}\label{sec:liter_semseg}

We first give a brief introduction to the dataset used for 3D scene understanding. For a comprehensive survey, please refer to \cite{guo2019deep} for more details.

In general, existing representative datasets can be roughly categorized into the following four subgroups based on the spatial coverage: \textbf{1) Object-level 3D models}. Early datasets are mainly focused on the recognition of individual objects, thereby usually composed of a collection of synthetic 3D CAD models. Representative datasets include the synthetic ModelNet \cite{3d_shapenets}, ShapeNet \cite{chang2015shapenet}, ShapePartNet \cite{ShapePartNet}, PartNet \cite{mo2019partnet} and the real-world ScanObjectNN \cite{scanobjectnn}. \textbf{2) Indoor scene-level 3D scans}. These datasets are usually acquired and further reconstructed by using commodity short-range depth scanners in indoor environments, including NYU3D \cite{NYU3D}, SUN RGB-D \cite{sunrgbd}, S3DIS \cite{2D-3D-S}, SceneNN \cite{scenenn}, Matterport3D \cite{Matterport3D} and ScanNet \cite{scannet}. Additionally, the SceneNet \cite{handa2015scenenet} and SceneNet RGB-D \cite{scenenetrgbd} dataset also provide large-scale photorealistic rendering of indoor synthetic layouts. \textbf{3) Outdoor roadway-level 3D point clouds.} Most of these datasets are driven by the increasing demand of autonomous driving application, and usually collected by using modern laser scanner systems, including static Terrestrial Laser Scanners (TLS) and Mobile Laser Scanners (MLS). Representative datasets include  the early Oakland \cite{oakland}, KITTI \cite{geiger2012we}, Sydney Urban Objects \cite{de2013unsupervised} and the recent Semantic3D \cite{Semantic3D}, Paris-Lille-3D \cite{NPM3D}, Argoverse \cite{chang2019argoverse}, SemanticKITTI \cite{behley2019semantickitti}, SemanticPOSS \cite{pan2020semanticposs}, Toronto-3D \cite{Toronto3D}, nuScenes \cite{caesar2020nuscenes}, A2D2 \cite{geyer2020a2d2}, CSPC-Dataset \cite{tong2020cspc}, Lyft dataset\footnote{\url{https://self-driving.lyft.com/level5/data/}} and Waymo dataset \cite{Waymo}. Additionally, synthetic datasets \cite{ros2016synthia,gaidon2016virtual} composed of realistic simulation of LiDAR point clouds are also included. \textbf{4) Urban-level aerial 3D point clouds.} These datasets are usually acquired by professional-grade airborne LiDAR systems, including the recent DublinCity \cite{zolanvari2019dublincity}, DALES \cite{varney2020dales} and LASDU \cite{ye2020lasdu}. Lacking the color information is the main limitation of these datasets, especially for the fine-grained semantic understanding of 3D scenarios. Interestingly, the very recent OpenGF \cite{qin2021opengf} dataset has started to investigate ultra-large-scale ground filtering datasets. However, this dataset mainly focuses on the task of ground extraction, instead of the fine-grained semantic understanding.

The recent Campus3D \cite{li2020campus3d} 3DOM \cite{3DOM}, and H3D \cite{H3D} are the most similar datasets to our SensatUrban datasets. They are also composed of large-scale photogrammetric 3D point clouds generated from high-resolution aerial images. However, our \nicknameData{} provides larger-scale 3D urban scenes with several times the number of points, as well as richer semantic annotations.

\subsection{Semantic Learning of 3D Scenes}
Thanks to the wide availability of various different 3D datasets, a large number of insightful research works have been presented and facilitated. The tremendous progress in semantic learning in turn greatly improved the best performance in several competitive leaderboards. Fundamentally, the semantic learning of 3D point clouds can be attributed to a representation learning problem, and the existing neural architectures can be roughly divided into the following three paradigms:

\textbf{1) Voxel-based approaches.} Early works \cite{le2018pointgrid, vvnet, tchapmi2017segcloud} usually voxelize point clouds into dense cubic grids, and then leverage the mature 3D CNN architectures to learn the semantics of each point. Although promising results have been achieved on several benchmarks, these techniques usually require cubically growing computation and memory with the input resolution. This severely limits the application of these methods on large-scale point clouds. To reduce the computational and memory cost, the sparse volumetric representation \cite{sparse, 4dMinkpwski, 2-S3Net} and point-voxel joint representation \cite{Point_voxel_cnn, e3d} are further introduced. Additionally, various different volumetric representations such as spherical voxels \cite{SPH3D}, cylindrical voxels \cite{cylinder3d} are also proposed to adapt to the data distribution of specific point clouds (\eg LiDAR).

\textbf{2) 2D projection-based methods}. Similarly, these pipelines \cite{rangenet++,lyu2020learning,salsanext,wu2018squeezeseg, wu2019squeezesegv2, xu2020squeezesegv3} leverage the well-developed 2D CNN frameworks to learn 3D semantics after projecting the point clouds onto 2D images. However, critical geometric information is very likely to be dropped in the 3D-2D projection (\eg the commonly-used birds-eye-view images), and therefore they are not suitable to learn the relatively small object categories within urban-scale scenarios.

\textbf{3) Point-based architectures} \cite{qi2017pointnet,qi2017pointnet++,li2018pointcnn,dgcnn,thomas2019kpconv,hu2019randla,wu2018pointconv,yan2020pointasnl,3PRNN,boulch2019generalizing,GACNet}. 
These methods directly operate on the unstructured point clouds, without relying on any explicit intermediate regular representation. This is achieved by using the simple shared MLPs to learn individual per-point features, and symmetrical aggregation functions to ensure permutation invariance \cite{qi2017pointnet}. In particular, PointNet++ \cite{qi2017pointnet++} is proposed to hierarchically learn the local features, DGCNN \cite{dgcnn} is introduced to model the topological structure through a graph architecture with Edge-Conv operation. A kernel point convolution \cite{thomas2019kpconv} is proposed to learn spatially correlation in unstructured point clouds.  \cite{hu2019randla} explore the efficient semantic learning of large-scale point clouds based on the point-based framework. Due to the simple implementation and straightforward architecture, this class of techniques has been widely investigated in several relevant tasks including 3D object detection \cite{zhou2018voxelnet, lang2019pointpillars} and instance segmentation \cite{3dbonet, jiang2020pointgroup}. However, it remains unclear whether the existing point pipelines can be well generalized to urban-scale point clouds. To this end, we build our SensatUrban dataset and investigate the unique challenges arising from the semantic understanding of urban-scale scenarios.

\section{Dataset Acquisition and Annotation}
\label{sec:dataset}
In this section, we first describe how we collect (Sec. \ref{sec:collect}) and reconstruct (Sec. \ref{sec:reconstruct}) the urban-scale 3D point clouds using UAV photogrammetry techniques, followed by the detailed procedures to label the dataset over several large urban areas in the UK (Sec. \ref{sec:annotation}).

\begin{figure}[thb]
\centering
\includegraphics[width=0.45\textwidth]{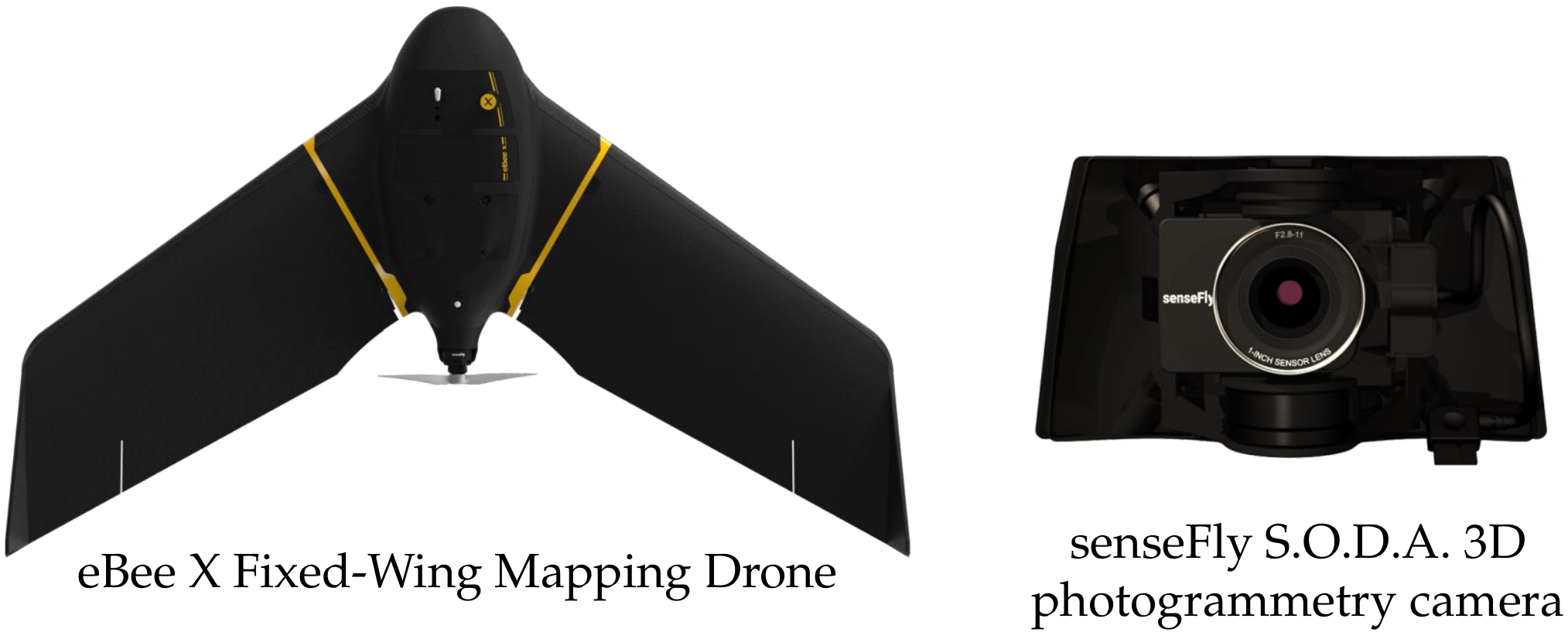}
\caption{The drones and cameras we used in the urban survey.}
\label{fig:drones}
\end{figure}

\subsection{Sequential Aerial Imagery Acquisition}
\label{sec:collect}
Considering the clear advantages of UAV photogrammetry over similar mapping techniques (such as LiDAR) in terms of cost, data quality, and practicality, we adopt a cost-effective fixed-wing mapping drone, Ebee X\footnote{https://www.sensefly.com/drone/ebee-x-fixed-wing-drone/}, equipped with a cutting-edge Sensefly S.O.D.A. camera, to stably capture high-resolution aerial image sequences, as shown in Figure \ref{fig:drones}. Note that, the camera has the ability to take both oblique and nadir photographs, ensuring that vertical surfaces are captured appropriately. The detailed specification of the camera can be found in Table \ref{tab:camera}.

\begin{figure}[t]
\centering
\includegraphics[width=0.5\textwidth]{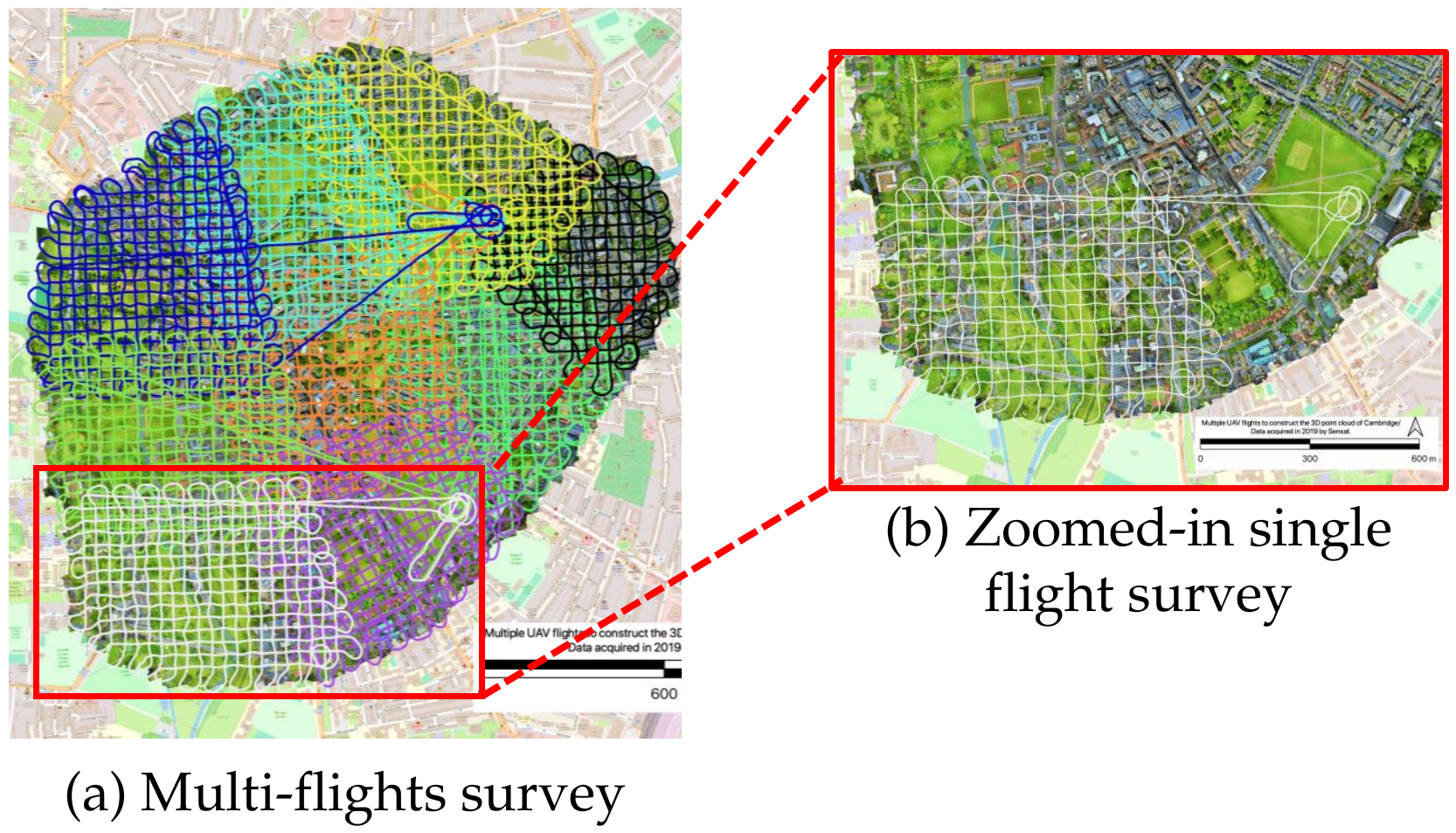}
\caption{An illustration of the survey in a region of Cambridge. A total of 9 flights were carried out together to cover the whole site. Different flight paths of UAVs are represented in lines with different colors. Note that, the drones fly in a grid fashion (\ie perpendicular flight) to capture more details of the facades of the urban environment. The circular path is the takeoff and landing pattern. }
\label{fig:survey}
\end{figure}

In order to fully and evenly cover the survey area, all flight paths are pre-planned in a grid fashion and automated by the flight control system (e-Motion). Several factors have been taken into consideration during the data collection workflow: the area covered, the flying permissions, the level of detail required, and the resolution needed, etc. In light of the limited battery capacity, multiple individual flights are applied in sequence to capture the whole site (each flight lasts between 40-50 minutes). For illustration, Figure \ref{fig:survey} shows the paths of the pre-planed multiple flights to cover the selected area in Cambridge city.

These multiple aerial image sequences can then be geo-referenced by Ground Control Points (GCPs) which can be measured by independent professional surveyors with high precision GNSS equipment. Alternatively, the Cambridge data is directly geo-referenced using a highly precise onboard Realtime Kinematic (RTK) GNSS and the final horizontal and vertical RMSEs are $\pm$50 mm and $\pm$75 mm, respectively (note they can be improved by introducing GCPs). As a comparison, the expected positioning accuracy of point clouds acquired by airborne LiDAR is around 5 to 10 cm, depending on the equipment quality, flying configuration, post-processing, etc. \cite{zhang2018patch}.  The resolution (point density) of our data depends on the number of input images and 3D reconstruction settings. Normally, photogrammetric point clouds are very dense from the process of dense image matching and so need to be subsampled. In our case, all points are subsampled to 2.5 cm, which is denser than most LiDAR data such as DALES \cite{varney2020dales}.

\begin{table}[t]
\centering
\caption{Detailed specifications of the camera (\ie Sensefly SODA 3D camera) used in our survey.}
\resizebox{0.45\textwidth}{!}{%
\begin{tabular}{rr}
\Xhline{2.0\arrayrulewidth}
                                             & Specification                                   \\
\Xhline{1.25\arrayrulewidth} 
Sensor size                                  & 1 inch                                       \\
RGB Lens                                     & F/2.8-11, 10.6 mm (35 mm equivalent: 29 mm)  \\
RGB Resolution                               & 5,472 x 3,648 px (3:2)                       \\
{\color[HTML]{212526} Exposure compensation} & ±2.0 (1/3 increments)                        \\
{\color[HTML]{212526} Shutter}               & Global Shutter 1/30 – 1/2000s                \\
{\color[HTML]{212526} White balance}         & Auto, sunny, cloudy, shady                   \\
{\color[HTML]{212526} ISO range}             & 125-6400                                     \\
{\color[HTML]{212526} RGB FOV}               & Total FOV: 154°, 64° optical, 90° mechanical \\
GNSS                                         & RTK/PPK                                      \\
\Xhline{2.0\arrayrulewidth}
\end{tabular}%
}
\label{tab:camera}
\end{table}

\begin{figure*}[t]
\centering
\includegraphics[width=1.0\textwidth]{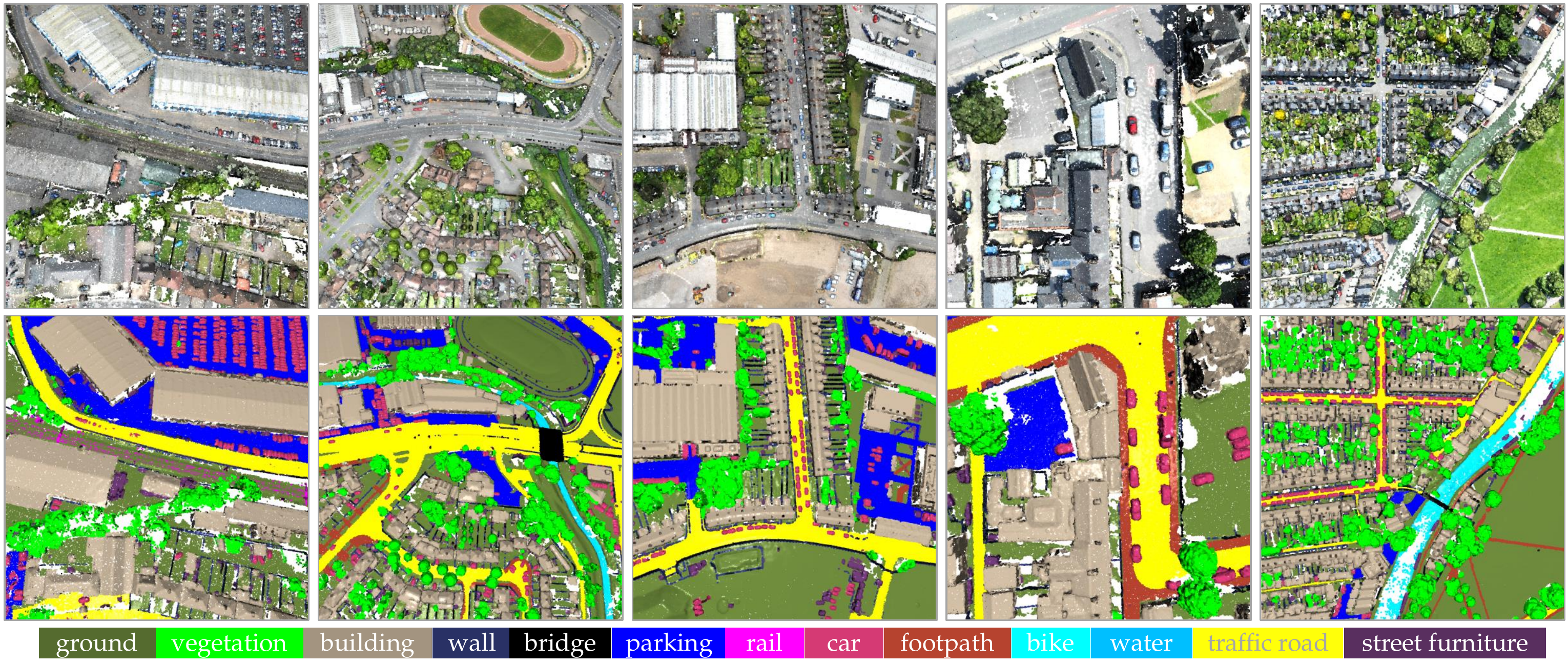}
\caption{ Visualization of example point cloud tiles in our \nicknameData{} dataset. \textbf{Top}: the raw point clouds. \textbf{Bottom}: the semantic annotations of corresponding point clouds. Points belonging to different semantic categories are displayed in different colors.}
\label{fig:annotation_examples}
\end{figure*}

\subsection{Urban-Scale 3D Point Clouds Reconstruction}
\label{sec:reconstruct}

Our SensatUrban dataset is reconstructed by using the well-established Structure-from-Motion with Multi-View Stereo  (SfM-MVS) techniques\cite{westoby2012structure} on the highly overlapped 2D aerial image sequences. The camera positions and orientation, and the scene geometry are first recovered simultaneously using a highly redundant iterative bundle adjustment, based on matched features extracted from overlapping offset images. The multi-view stereo image matching technique is then applied to reconstruct dense and coloured 3D point clouds.

In this paper, we use the off-the-shelf software Pix4D\footnote{https://www.pix4d.com/} to generate the 3D point clouds and orthomosasics. The final outputs of the survey include reconstructed 3D point clouds, 2D orthomosaic images, and 2.5D Digital Surface Model (DSM). In this work, we focus on the 3D point clouds, while the byproduct orthomosaics are only used for visualization purposes. Specifically, we feed all the captured sequential images to Pix4D to generate the 3D point clouds of each region, including the urban area on the periphery of Birmingham, the urban region adjacent to the city centre of Cambridge, and the central area of York. The statistics of the final output point clouds are summarized in Table \ref{tab:reconstructed_pc}.

\begin{table}[thb]
\centering
\caption{Statistics of the reconstructed 3D point clouds in different cities. The area of the surveyed region and the number of generated points are reported.}
\resizebox{0.38\textwidth}{!}{%
\begin{tabular}{ccr}
\Xhline{2.0\arrayrulewidth}
City & Area ($km^2$) & Number of Points \\ 
\Xhline{1.25\arrayrulewidth} 
Birmingham & 1.2 & 569,147,075 \\
Cambridge & 3.2 & 2,278,514,725 \\
York & 3.2 & 904,155,619 \\
\textbf{Total} & 7.6 & 3,751,817,419 \\ 
\Xhline{2.0\arrayrulewidth}
\end{tabular}%
}
\label{tab:reconstructed_pc}
\end{table}

\subsection{Point-wise Semantic Annotations}
\label{sec:annotation}
To provide fine-grained information of our dataset for subsequent tasks, we further enrich the reconstructed 3D point clouds with point-wise semantic annotations. However, it is non-trivial and particularly important to decide the categories of interest before manual annotation. In this paper, we specify the semantic categories based on the following three principles: 1) Each annotated category should be of interest to social or commercial purposes, such as asset management \cite{assest_management}, automated structural damage assessment \cite{damage_assessment}, and urban planning \cite{urban_planning}, etc. 2) Each category should have a clear and unambiguous semantic meaning. 3) Different categories should have significant variance in terms of geometric structure or appearance.

Based on these three criteria, we first labeled the point cloud as highly detailed 31 categories via off-the-shelf point cloud labeling tools (\ie CloudCompare), including fine-grained urban elements such as \textit{benches}, \textit{bollards}, \textit{road signs}, \textit{traffic lights}, etc. Considering the scarcity of data points in certain categories, we merged some similar categories together and finally identified the below 13 semantic classes for all the 3D points in Birmingham and Cambridge. The detailed definition of the semantic categories are shown in Table \ref{tab:class_denifinitions}. The points in York remain unlabeled, but made available for possible pre-training in semi-supervised schemes. To ensure the annotation quality, all points are annotated independently by two professional operators in the first round. This is followed by cross-validation in the second round. We also give timely and regular feedback to annotators to address potential issues. All discrepancies in the annotation are carefully addressed, greatly reducing the biases from operators and keeping the annotations consistent and high quality. It takes around 600 working hours to label the entire dataset, and there are no unassigned points discarded in the process. Figure \ref{fig:annotation_examples} shows examples of our annotations. Table \ref{tab:Overview-dataset-1} compares the statistics of our \nicknameData{} with a number of existing 3D datasets.

\begin{table}[thb]
\centering
\caption{Class definitions and ordering of our \nicknameData{} dataset.}
\label{tab:class_denifinitions}
\resizebox{0.5\textwidth}{!}{%
\begin{tabular}{ccc}
\Xhline{2.0\arrayrulewidth}
Class number & Class name & Definition \\ 
\Xhline{1.25\arrayrulewidth}
1 & Ground & impervious surfaces, grass, terrain \\
2 & Vegetation & trees, shrubs, hedges, bushes \\
3 & Building & commercial / residential buildings \\
4 & Wall & fence, highway barriers, walls \\
5 & Bridge & road bridges \\
6 & Parking & parking lots \\
7 & Rail & railroad tracks \\
8 & Traffic Road & main streets, highways, drivable areas \\
9 & Street Furniture & benches, poles, lights \\
10 & Car & cars, trucks, jeeps, SUVs, HGVs \\
11 & Footpath & walkway, alley \\
12 & Bike & bikes / bicyclists \\
13 & Water & rivers / water canals \\ 
\Xhline{2.0\arrayrulewidth}
\end{tabular}%
}
\end{table}

Note that, our SensatUrban dataset not only incorporates common categories such as \textit{ground}, \textit{building}, and \textit{vegetation}, but also involves several new categories that were not included in the previous urban-scale point cloud datasets \cite{varney2020dales, zolanvari2019dublincity}, such as \textit{rail}, \textit{bridge}, and \textit{water}. In particular, these categories are derived by discussing with the industry professionals, and are particularly important for urban planning and infrastructure mapping.

The SensatUrban dataset has been made publicly available\footnote{http://point-cloud-analysis.cs.ox.ac.uk}, all point clouds and the point-wise ground-truth label of the training set are provided for network training, and an online hidden test set\footnote{https://competitions.codalab.org/competitions/31519} is used to evaluate the final segmentation performance. To prevent overfitting on the test set, the maximum number of submissions is also limited.

\begin{figure*}[t]
\centering
\includegraphics[width=1.0\textwidth]{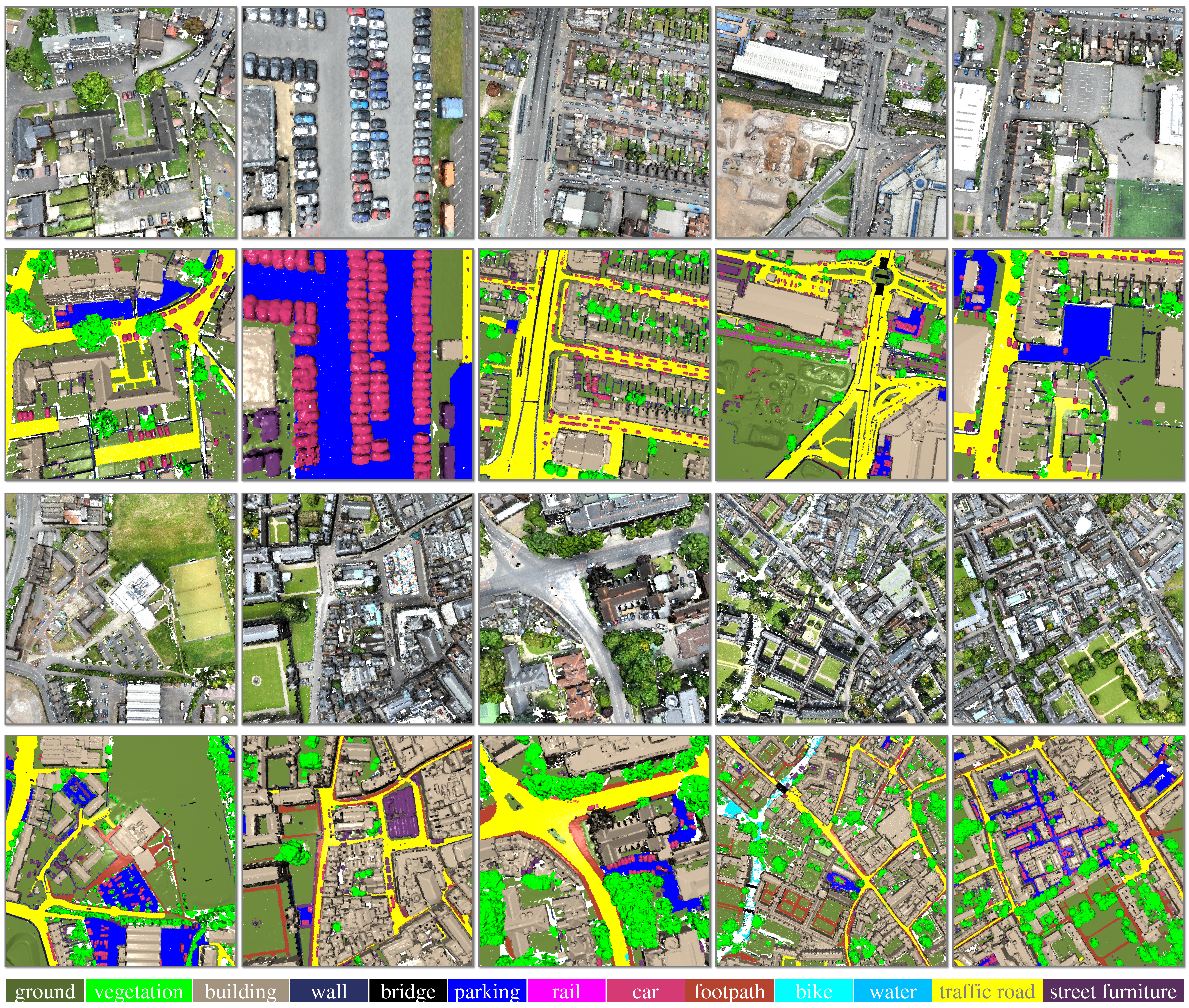}
\caption{Additional examples of our \nicknameData{} dataset. Different semantic classes are labeled by different colors. The top two rows are point clouds collected from Birmingham, and the bottom two rows are point clouds acquired from Cambridge.}
\label{fig:extra_annotation_examples}
\end{figure*}

\begin{figure*}[htb]
\centering
\includegraphics[width=1.0\textwidth]{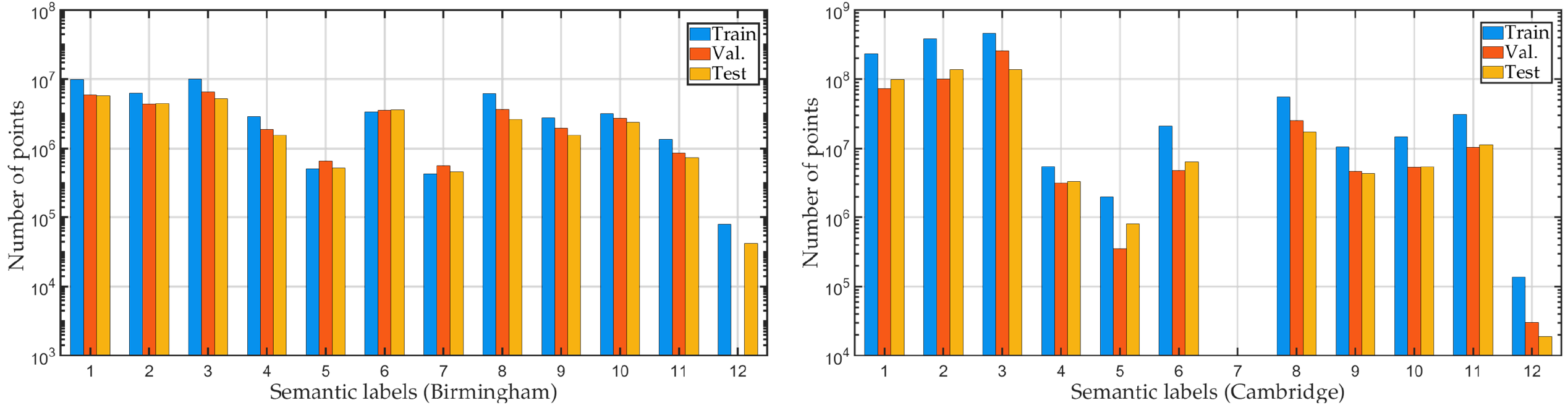}
\caption{Statistics of our \nicknameData{} dataset. The number of points in different semantic categories is reported. Please note that the vertical axis is on the logarithmic scale. Additionally, there are no points annotated as \textit{rail} in Cambridge. }
\label{fig:statistics}
\end{figure*}

\section{Benchmarks}
\label{sec:benchmarks}
\begin{table*}[thb]
\centering
\caption{Evaluation of selected baselines on our \nicknameData{} benchmark. We evaluate on 13 predefined semantic categories using the Overall Accuracy (OA, \%), mean class Accuracy (mAcc, \%), mean IoU (mIoU, \%), and per-class IoU (\%). \label{tab:benchmarks}}
\resizebox{\textwidth}{!}{%
\begin{tabular}{l|cccccccccccccccc}
\Xhline{2.0\arrayrulewidth}
 & \rotatebox{90}{OA(\%)} &  \rotatebox{90}{mAcc(\%)} & \rotatebox{90}{\textbf{mIoU(\%)}} & \rotatebox{90}{ground} & \rotatebox{90}{veg.} & \rotatebox{90}{building} & \rotatebox{90}{wall} & \rotatebox{90}{bridge} & \rotatebox{90}{parking} & \rotatebox{90}{rail} & \rotatebox{90}{traffic.} & \rotatebox{90}{street.} & \rotatebox{90}{car} & \rotatebox{90}{footpath} & \rotatebox{90}{bike} & \rotatebox{90}{water} \\
\Xhline{1.25\arrayrulewidth} 
PointNet \cite{qi2017pointnet} & 80.78 & 30.32 & 23.71 & 67.96 & 89.52 & 80.05 & 0.00 & 0.00 & 3.95 & 0.00 & 31.55 & 0.00 & 35.14 & 0.00 & 0.00 & 0.00\\
PointNet++ \cite{qi2017pointnet++} & 84.30 & 39.97 & 32.92 & 72.46 & 94.24 & 84.77 & 2.72 & 2.09 & 25.79 & 0.00 & 31.54 & 11.42 & 38.84 & 7.12 & 0.00 & 56.93\\
TangentConv \cite{tangentconv} &76.97 & 43.71 & 33.30 & 71.54 & 91.38 & 75.90 & 35.22 & 0.00 &  45.34 & 0.00 & 26.69 & 19.24 & 67.58 & 0.01 & 0.00 & 0.00 \\
SPGraph \cite{landrieu2018large} & 85.27 & 44.39 & 37.29 & 69.93 & 94.55 & 88.87 & 32.83 & 12.58 & 15.77 & \textbf{15.48} & 30.63 & 22.96 & 56.42 & 0.54 & 0.00 & 44.24\\
SparseConv \cite{sparse} & 88.66 & 63.28 & 42.66 & 74.10 & 97.90 & 94.20 & 63.30 & 7.50 & 24.20 & 0.00 & 30.10 & 34.00 & 74.40 & 0.00 & 0.00 & 54.80\\
KPConv \cite{thomas2019kpconv} & \textbf{93.20} & 63.76 & \textbf{57.58} & \textbf{87.10} & \textbf{98.91} & \textbf{95.33} & \textbf{74.40} & 28.69 & 41.38 & 0.00 & 55.99 & \textbf{54.43} & \textbf{85.67} & \textbf{40.39} & 0.00 & \textbf{86.30}\\
RandLA-Net \cite{hu2019randla} & 89.78 & \textbf{69.64} & 52.69 & 80.11 & 98.07 & 91.58 & 48.88 & \textbf{40.75} & \textbf{51.62} & 0.00 & \textbf{56.67} & 33.23 & 80.14 & 32.63 & 0.00 & 71.31\\
\Xhline{2.0\arrayrulewidth}
\end{tabular}}
\end{table*}

\subsection{Statistics of Train/Val/Test Split}
Based on the proposed SensatUrban dataset, we further set up a benchmark to evaluate the performance of existing state-of-the-art segmentation methods. Notably, we first follow DALES \cite{varney2020dales} to divide the urban-scale point clouds into similarly sized small tiles (without overlap), so that existing methods can be trained and tested on modern GPUs. Specifically, the urban point clouds collected in Birmingham have been split into 14 tiles, and the Cambridge point clouds are similarly divided into 29 tiles in total. Note that, each tile is approximately 400$\times$400 square meters. We also report the detailed statistics of the training/validation/testing subsets in both Birmingham and Cambridge in Figure \ref{fig:statistics}. It can be seen that the number of points belongs to different semantic categories varies greatly. For example, the dominant three semantic categories, \textit{i.e.}, \textit{ground / building / vegetation},  together account for more than 50\% of the total points. However, the minor yet important two categories (\textit{e.g.}, \textit{bike / rail}) only account for 0.025\% of the total points. This clearly shows that the class distribution of our \nicknameData{} dataset is extremely imbalanced, which is in correspondence with the long-tailed distribution of real data. As described in Sec. \ref{sec:benchmarks}, the imbalanced distribution nature of our dataset also poses great challenges in generalizing the existing segmentation approaches.

\subsection{Representative Baselines}

To comprehensively evaluate the performance of existing point cloud segmentation pipelines on urban-scale point clouds, we carefully select 7 representative approaches, which cover the three mainstream paradigms as discussed in Section \ref{sec:liter_semseg}, as solid baselines in our SensatUrban benchmark. A short summary of these baselines are as follows:
\begin{itemize}[leftmargin=*]
\item SparseConvNet \cite{sparse}. A strong baseline that introduces the submanifold sparse convolutional networks for efficient semantic segmentation of 3D point clouds. This method and its follow-up works \cite{4dMinkpwski, han2020occuseg} lead the ScanNet benchmark.
\item TangentConv \cite{tangentconv}. This method introduces tangent convolution, which operates directly on surface geometry, for large-scale point clouds processing. In particular, the 3D point clouds are first projected as tangent images, followed by 2D convolutional networks.
\item PointNet \cite{qi2017pointnet}. This is the pioneering work for directly operating on orderless point clouds by using shared MLPs and symmetrical max-pooling aggregation.
\item PointNet++ \cite{qi2017pointnet++}. This is the follow-up work of PointNet. It introduced multi-scale/resolution grouping to extract local geometrical patterns, and farthest point sampling to reduce memory and computational cost.
\item KPConv \cite{thomas2019kpconv}. This approach presents a  powerful kernel point convolution to learn spatially correlation from unstructured 3D point clouds.  A set of rigid or deformable kernel points are placed to learn varying local geometries. It has achieved state-of-the-art performance on the aerial DALES dataset \cite{varney2020dales}.
\item SPGraph \cite{landrieu2018large}. This is one of the first learning-based frameworks that capable of processing large-scale point clouds with millions of points. The pipeline composed of geometrically homogeneous partition, followed by superpoint graph construction and contextual segmentation. It is one of the top-performing approaches on the Semantic3D dataset.
\item RandLA-Net \cite{hu2019randla}. It is one of the latest works for efficient semantic understanding of large-scale point clouds. The computational and memory-efficient random sampling, and the hierarchical local feature aggregation are the key to the great performance of this method. It also achieves leading performance on the Semantic3D leaderboard \cite{Semantic3D}.
\end{itemize}

\subsection{Evaluation Metrics}
Similar to most of the existing 3D point cloud benchmarks \cite{Semantic3D,behley2019semantickitti,2D-3D-S}, \textit{Overall Accuracy} (OA), \textit{mean class Accuracy} (mAcc), and \textit{mean Intersection-over-Union} (mIoU) are adopted as the primary evaluation criteria of our \nicknameData{} benchmark. The detailed score of each metric is calculated as follows:

\begin{ceqn}
\begin{align}
\mathrm{OA}=\frac{\mathrm{TP}}{\mathrm{TP}+\mathrm{FP}+\mathrm{FN}+\mathrm{TN}}
\end{align}
\end{ceqn}

\begin{ceqn}
\begin{align}
\mathrm{mAcc}=\frac{1}{C}\sum_{c=1}^{C}\frac{\mathrm{TP}_{c}}{\mathrm{TP}_{c}+\mathrm{FN}_{c}}
\end{align}
\end{ceqn}

\begin{ceqn}
\begin{align}
\mathrm{mIoU}=\frac{1}{C}\sum_{c=1}^{C}\frac{\mathrm{TP}_{c}}{\mathrm{TP}_{c}+\mathrm{FP}_{c}+\mathrm{FN}_{c}}
\end{align}
\end{ceqn}
where $\mathrm{TP}$, $\mathrm{TN}$, $\mathrm{FP}$, $\mathrm{FN}$ denote the true-positive, true-negative, false positive, and false negative, separately. $C$ is the total number of semantic categories. Note that, these scores can also be calculated based on the confusion matrix.

\subsection{Benchmark Results}
We then evaluate the performance of the aforementioned baselines on our urban-scale SensatUrban dataset. The quantitative results including the per-class IoU scores are reported in Table \ref{tab:benchmarks}. Note that, we faithfully follow the experimental settings and the publicly available implementation provided by each baseline in their original manuscript. All methods are trained on the same training split for a fair comparison.

Not surprisingly, the performance of all baselines shows varying degrees of degradation, compared with that achieved in other similar aerial point cloud datasets \cite{varney2020dales}. Specifically, the recent KPConv \cite{thomas2019kpconv} shows the best mIoU performance, but only with a mIoU score of 57.58\%, which is still far from satisfactory in practice. In particular, several infrastructure-oriented semantic categories such as \textit{rail}, \textit{footpath}, and \textit{bridge} are poorly segmented. Additionally, we also noticed that the category \textit{bike} is completely misclassified by all baselines. In general, all baselines are more likely to achieve better segmentation performance in categories with simple geometrical structure and dominant proportion, such as \textit{ground}, \textit{vegetation}, \textit{building} and \textit{car}, while achieving relatively limited performance on categories such as \textit{wall}, \textit{bridge}, and \textit{water}. Additionally, different baselines have vastly different performances in individual semantic categories, without a clear leader. Overall, there remain several particular challenges for selected baselines to achieve satisfactory segmentation performance in the proposed city-scale SensatUrban dataset. Motivated by this, we then dive deep into the challenges that arise from our new urban-scale dataset.

\section{Challenges}\label{sec:challenges}

In this section, we further analyze the key challenges to generalize existing deep segmentation models to urban-scale photogrammetry point clouds. In particular, we first identify several unique challenges from the perspective of dataset characteristics. Next, we further explore the potential solutions and perform specific experiments to verify the effectiveness. Note that, this paper is not aiming to introduce specific new algorithms to solve all these challenges, but hopes to point out unresolved issues and provide in-depth analysis and insights, eventually stimulate the development of fine-grained urban-scale point cloud understanding.

\begin{table*}[thb]
\centering
\caption{Semantic segmentation results achieved by selected baselines \cite{hu2019randla, qi2017pointnet} with different input preparation steps. Note that, the performance of all baselines is evaluated on the original point clouds, instead of a downsampled point cloud. \label{tab:input-preparation}}
\resizebox{\textwidth}{!}{%
\begin{tabular}{rcccccccccccccccccc}
\Xhline{2.0\arrayrulewidth}
 & Sampling & Input sets & \rotatebox{90}{OA(\%)} &  \rotatebox{90}{mAcc(\%)} & \rotatebox{90}{\textbf{mIoU(\%)}} & \rotatebox{90}{ground} & \rotatebox{90}{veg.} & \rotatebox{90}{building} & \rotatebox{90}{wall} & \rotatebox{90}{bridge} & \rotatebox{90}{parking} & \rotatebox{90}{rail} & \rotatebox{90}{traffic.} & \rotatebox{90}{street.} & \rotatebox{90}{car} & \rotatebox{90}{footpath} & \rotatebox{90}{bike} & \rotatebox{90}{water} \\
\Xhline{1.25\arrayrulewidth} 
PointNet & Grid & Constant Number& \textbf{90.57} & \textbf{56.30} & \textbf{49.69} & \textbf{83.55} & \textbf{97.67} & 90.66 & \textbf{22.56} & \textbf{43.54} & 40.35 & 9.29 & \textbf{50.74} & \textbf{29.58} & 68.24 & \textbf{29.27} & 0.00 & \textbf{80.55}\tabularnewline
PointNet & Grid & Constant Volume & 88.27 & 49.80 & 42.44 & 80.20 & 96.43 & 87.88 & 8.45 & 35.14 & 32.52 & 0.00 & 43.03 & 19.26 & 54.66 & 18.26 & 0.00 & 75.87\tabularnewline
PointNet & Random & Constant Number& 90.34 & 55.17 & 48.49 & 83.47 & 97.51 & \textbf{90.89} & 18.55 & 33.31 & \textbf{42.82} & \textbf{11.85} & 47.95 & 26.83 & \textbf{68.37} & 29.12 & 0.00 & 79.71\tabularnewline
PointNet & Random & Constant Volume & 88.09 & 48.45 & 41.68 & 79.82 & 96.24 & 87.64 & 5.69 & 27.70 & 34.98 & 0.00 & 42.85 & 13.81 & 54.29 & 20.64 & 0.00 & 78.24\tabularnewline
\hline 
RandLA-Net & Grid & Constant Number& \textbf{91.55} & \textbf{74.87} & \textbf{58.64} & \textbf{82.99} & \textbf{98.43} & \textbf{93.41} & \textbf{57.43} & \textbf{49.47} & \textbf{55.12} & \textbf{27.33} & \textbf{60.65} & \textbf{39.43} & \textbf{84.57} & \textbf{39.48} & 0.00 & 73.97\tabularnewline
RandLA-Net & Grid & Constant Volume & 88.11 & 64.91 & 49.18 & 78.18 & 97.92 & 90.87 & 45.02 & 30.89 & 35.82 & 0.00 & 45.73 & 31.96 & 77.78 & 29.90 & 0.00 & 75.30\tabularnewline
RandLA-Net & Random & Constant Number& 91.14 & 74.14 & 57.55 & 82.25 & 98.33 & 92.37 & 54.20 & 43.10 & 54.74 & 25.02 & 60.40 & 39.17 & 82.77 & 37.59 & 0.00 & \textbf{78.25}\tabularnewline
RandLA-Net & Random & Constant Volume & 88.37 & 60.84 & 47.27 & 81.16 & 97.52 & 90.45 & 44.75 & 16.36 & 37.18 & 0.00 & 4219 & 26.28 & 76.76 & 30.46 & 0.00 & 71.39\tabularnewline
\Xhline{2.0\arrayrulewidth} 
\end{tabular}}
\end{table*}

\subsection{Data Preparation}
Considering the limited memory of modern GPUs, it is infeasible and unrealistic to directly accommodate and process urban-scale point clouds with billions of points in practice. As a result, the original point cloud data are usually partitioned into small pieces or downsampled before feed into existing neural architectures, so as to find a trade-off between the computational efficiency and segmentation accuracy. 

In particular, the early works including PointNet \cite{qi2017pointnet}, PointNet++ \cite{qi2017pointnet++}, and their variants usually first divide the large point clouds into equally-sized small blocks with partial overlap (\textit{e.g.}, 1m$\times$1m blocks in the S3DIS dataset \cite{2D-3D-S}). However, the final segmentation performance is highly-sensitive to the input block size. Large blocks with massive points lead to an unaffordable GPU memory cost, while small blocks inevitably break the objects' geometrical structure. Recent works such as KPConv \cite{thomas2019kpconv} and RandLA-Net \cite{hu2019randla} resort to grid or random down-sampling at the beginning to reduce the total amount of points. Additionally, several other works \cite{3PRNN} applied different partitioning or downsampling steps to preprocess the raw point clouds. Overall, various data preparation steps are intensively-involved in existing neural pipelines, but there are still no standard and principled preparation steps in literature, not to mention comprehensive evaluation and analysis.

To further investigate the impact of different data preparations on the final segmentation performance, we standardized a unified two-step data preprocessing framework. The detailed descriptions are as follows:

\begin{itemize}[leftmargin=*]
\item Step 1. Reducing the redundant points in the original point clouds through downsampling. This can be achieved by using 1) random downsampling \cite{hu2019randla} or 2) grid-downsampling \cite{thomas2019kpconv}. In particular, random downsampling has superior computational and memory efficiency, while grid downsampling is robust to varying point densities. Both methods can significantly reduce the total amount of points.
\item Step 2. Iteratively feeding mini-batches of point subsets into the network. This can be achieved by first constructing efficient space partitioning data structures such as a KDTree, and then either query 1) constant-number point subsets or 2) constant-volume point subsets from specific regions. In particular,  constant-number input sets are usually obtained by querying a fixed number of neighboring points with regard to a specific point \cite{hu2019randla}, while the constant-volume input sets are achieved by cropping fixed-size point cloud chunks (\eg cubes, spheres) \cite{qi2017pointnet,qi2017pointnet++} centered on a specific point. Note that, the query points are random initialized and dynamically updated as in \cite{thomas2019kpconv}.

\end{itemize}

To evaluate the impact of 4 different combinations of both Step 1 and Step 2 on the segmentation performance, we select two representative approaches PointNet \cite{qi2017pointnet} and RandLA-Net \cite{hu2019randla} as the baselines. For a fair comparison, we set the grid size for grid downsampling as 0.2m, while the downsampling ratio is set to 1/10, so as to keep similar number of points after downsampling operation. For constant-number inputs, we implement this by using a prebuilt KDTree to query a fixed number of neighboring points of the center point as inputs. For constant volume inputs, we first crop a fixed-size volume (\eg 8m$\times$8m block) around the center point, followed by random down(up)-sampling to align the number of different input sets.

\textbf{Analysis.} 
Table \ref{tab:input-preparation} reports the quantitative semantic segmentation scores achieved by baseline approaches with different input preparations. Table \ref{tab:downsample} shows the number of points left after downsampling, and the detailed time used for downsampling. It can be seen from the results that:
\begin{itemize}[leftmargin=*]
\item Both two baseline approaches consistently show better performance when adopting constant number input sets, compared with corresponding variants which using constant volume input sets.
\item Both PointNet and RandLA-Net show slightly better segmentation performance when using grid downsampling at the very beginning, compared with the counterpart using random downsampling. However, the total time consumption for grid downsampling is significantly large than using random downsampling (129s vs. 1107s) when evaluated on the same hardware configuration with an Intel Core™ i9-10900X CPU and an NVIDIA RTX 3090 GPU.
\end{itemize}

\begin{table}[t]
\centering
\caption{Comparison of the grid downsampling and random downsampling at the beginning of data preparation.}
\label{tab:downsample}
\resizebox{0.42\textwidth}{!}{%
\begin{tabular}{rcc}
\Xhline{2.0\arrayrulewidth}
\multicolumn{1}{c}{} & \begin{tabular}[c]{@{}c@{}}Number of points\\ after sampling\end{tabular} & \begin{tabular}[c]{@{}c@{}}Time \\ consumption(s)\end{tabular} \\
\toprule[0.5pt]
Grid downsample & 220,671,929 & 1107 \\
Random downsample & 270,573,783 & 129 \\
\Xhline{2.0\arrayrulewidth}
\end{tabular}%
}
\end{table}
To summarize, our experiments demonstrate the importance of data preparation for the semantic segmentation performance. Although this issue has been overlooked by the community for a long time, we show that the same network architecture can bring up to 10\% performance gap, when equipped with different data preparation steps. Therefore, it is desirable and encouraged to further investigate the effective data preparation schemes, especially for our urban-scale point cloud datasets.

\begin{table*}[thb]
\centering
\caption{Evaluation of semantic segmentation performance of five selected baselines on our \nicknameData{} dataset with/without the usage of color information.\label{tab:color}}
\resizebox{\textwidth}{!}{%
\begin{tabular}{lcccccccccccccccc}
\Xhline{2.0\arrayrulewidth}
 & \rotatebox{90}{OA(\%)} &  \rotatebox{90}{mAcc(\%)} & \rotatebox{90}{\textbf{mIoU(\%)}} & \rotatebox{90}{ground} & \rotatebox{90}{veg.} & \rotatebox{90}{building} & \rotatebox{90}{wall} & \rotatebox{90}{bridge} & \rotatebox{90}{parking} & \rotatebox{90}{rail} & \rotatebox{90}{traffic.} & \rotatebox{90}{street.} & \rotatebox{90}{car} & \rotatebox{90}{footpath} & \rotatebox{90}{bike} & \rotatebox{90}{water} \\
\Xhline{1.25\arrayrulewidth} 
PointNet \cite{qi2017pointnet} (w/o RGB) & 83.50 & 33.52 & 28.85 & 67.35 & 92.66 & 84.72 & 16.02 & 0.00 & 13.65 & 2.68 & 17.09 & 0.33 & 54.54 & 0.00 & 0.00 & 26.04\\
PointNet \cite{qi2017pointnet} (w/ RGB) & 90.57 & 56.30 & 49.69 & 83.55 & 97.67 & 90.66 & 22.56 & 43.54 & 40.35 & 9.29 & 50.74 & 29.58 & 68.24 & 29.27 & 0.00 & 80.55\\
\midrule[0.5pt]
PointNet++ \cite{qi2017pointnet++} (w/o RGB) & 90.85 & 56.94 & 50.71 & 79.05 & 98.37 & 94.22 & 66.76 & 39.74 & 37.51 & 0.00 & 51.53 & 38.82 & 81.71 & 5.80 & 0.00 & 65.68\\
PointNet++ \cite{qi2017pointnet++} (w RGB) & 93.10 & 64.96 & 58.13 & 86.38 & 98.76 & 94.72 & 65.91 & 50.41 & 50.53 & 0.00 & 58.40 & 46.95 & 82.31 & 38.40 & 0.00 & \textbf{82.88}\\
\midrule[0.5pt]
SPGraph \cite{landrieu2018large} (w/o RGB) & 84.81 & 42.12 & 35.29 & 69.60 & 94.18 & 88.15 & 34.55 & 20.53 & 15.83 & 16.34 & 31.44 & 10.54 & 55.01 & 0.98 & 0.00 & 21.57\\
SPGraph \cite{landrieu2018large} (w RGB) & 85.27 & 44.39 & 37.29 & 69.93 & 94.55 & 88.87 & 32.83 & 12.58 & 15.77 & 15.48 & 30.63 & 22.96 & 56.42 & 0.54 & 0.00 & 44.24\\
\midrule[0.5pt]
KPConv \cite{thomas2019kpconv} (w/o RGB) & 91.47 & 57.43 & 51.79 & 80.43 & 98.82 & 94.93 & 74.17 & 44.53 & 32.11 & 0.00 & 54.32 & 37.83 & 84.88 & 14.48 & 0.00 & 56.79\\
KPConv \cite{thomas2019kpconv} (w RGB) & \textbf{93.92} & 71.44 & \textbf{64.50} & \textbf{87.04} & \textbf{99.01} & \textbf{96.31} & \textbf{77.73} & \textbf{58.87} & 49.88 & \textbf{37.84} & \textbf{62.74} & \textbf{56.60} & \textbf{86.55} & \textbf{44.86} & 0.00 & 81.01\\
\midrule[0.5pt]
RandLA-Net \cite{hu2019randla} (w/o RGB) & 88.90 & 67.96 & 51.53 & 77.30 & 97.92 & 91.24 & 51.94 & 47.46 & 45.04 & 9.71
& 49.79 & 34.21 & 79.97 & 21.13 & 0.00 & 64.18\\
RandLA-Net \cite{hu2019randla} (w RGB) & 91.24 & \textbf{74.68} & 58.14 & 82.23 & 98.39 & 92.69 & 56.62 & 49.00 & \textbf{54.19} & 25.10 & 60.98 & 38.69 & 83.42 & 38.74 & 0.00 & 75.80\\
\Xhline{2.0\arrayrulewidth}
\end{tabular}}
\end{table*}

\begin{table*}[thb]
\centering
\caption{Evaluation of semantic segmentation performance of  PointNet \cite{qi2017pointnet} and RandLA-Net \cite{hu2019randla} with different loss functions. \label{tab:loss}}
\resizebox{\textwidth}{!}{%
\begin{tabular}{lcccccccccccccccc}
\Xhline{2.0\arrayrulewidth}
 & \rotatebox{90}{OA(\%)} &  \rotatebox{90}{mAcc(\%)} & \rotatebox{90}{\textbf{mIoU(\%)}} & \rotatebox{90}{ground} & \rotatebox{90}{veg.} & \rotatebox{90}{building} & \rotatebox{90}{wall} & \rotatebox{90}{bridge} & \rotatebox{90}{parking} & \rotatebox{90}{rail} & \rotatebox{90}{traffic.} & \rotatebox{90}{street.} & \rotatebox{90}{car} & \rotatebox{90}{footpath} & \rotatebox{90}{bike} & \rotatebox{90}{water} \\
\Xhline{1.25\arrayrulewidth} 
PointNet+ce & \textbf{90.57} & 56.30 & 49.69 & \textbf{83.55} & \textbf{97.67} & \textbf{90.66} & 22.56 & 43.54 & 40.35 & 9.29 & \textbf{50.74} & 29.58 & \textbf{68.24} & 29.27 & 0.00 & \textbf{80.55}\\
PointNet+wce \cite{hu2019randla} & 88.13 & \textbf{68.05} & 51.24 & 81.01 & 97.12 & 87.87 & 24.46 & 45.76 & 47.78 & \textbf{34.93} & 49.82 & 29.58 & 61.28 & 31.78 & 0.00 & 74.67\\
PointNet+wce+sqrt \cite{salsanet} & 89.72 & 67.97 & 52.35 & 82.87 & 97.33 & 90.42 & 28.32 & 44.94 & 48.39 & 32.07 & 49.58 & \textbf{32.63} & 65.11 & 32.59 & \textbf{2.60} & 73.71\\
PointNet+lovas \cite{berman2018lovasz} & 89.58 & 67.50 & \textbf{52.53} & 82.74 & 97.27 & 90.28 & 28.11 & 43.89 & \textbf{48.53} & 33.58 & 49.68 & 32.21 & 64.01 & \textbf{33.05} & 1.46 & 78.13\\
PointNet+focal \cite{focal_loss} & 89.46 & 67.33 & 52.37 & 82.47 & 97.34 & 90.25 & \textbf{28.36} & \textbf{51.87} & 46.40 & 30.50 & 48.62 & 32.43 & 65.00 & 32.23 & 1.21 & 74.10\\
\midrule[0.5pt] 
RandLA-Net+ce & \textbf{93.10} & 64.30 & 57.77 & \textbf{85.39} & \textbf{98.63} & \textbf{95.40} & 62.55 & 54.85 & 56.49 & 0.00 & 58.13 & \textbf{45.90} & 82.24 & 30.68 & 0.00 & 80.70\\
RandLA-Net+wce \cite{hu2019randla} & 91.24 & 74.68 & 58.14 & 82.23 & 98.39 & 92.69 & 56.62 & 49.00 & 54.19 & 25.10 & \textbf{60.98} & 38.69 & 83.42 & 38.74 & 0.00 & 75.80\\
RandLA-Net+wce+sqrt \cite{salsanet} & 92.51 & \textbf{79.92} & \textbf{62.80} & 84.94 & 98.47 & 95.07 & 59.01 & \textbf{62.18} & 56.76 & 28.96 & 57.36 & 44.47 & \textbf{84.67} & 41.67 & \textbf{24.31} & 78.49\\
RandLA-Net+lovas \cite{berman2018lovasz} & 92.56 & 76.99 & 61.51 & 84.92 & 98.55 & 94.64 & \textbf{63.17} & 52.37 & 55.43 & \textbf{36.37} & 59.35 & 45.79 & 84.28 & 41.24 & 2.66 & \textbf{80.89}\\
RandLA-Net+focal \cite{focal_loss} & 92.49 & 77.26 & 60.41 & 85.03 & 98.38 & 94.74 & 59.49 & 58.70 & \textbf{57.11} & 25.97 & 58.19 & 42.74 & 82.26 & \textbf{42.00} & 2.71 & 77.97\\
\Xhline{2.0\arrayrulewidth}
\end{tabular}}
\end{table*}

\begin{figure*}[thb]
\centering
\includegraphics[width=1.0\textwidth]{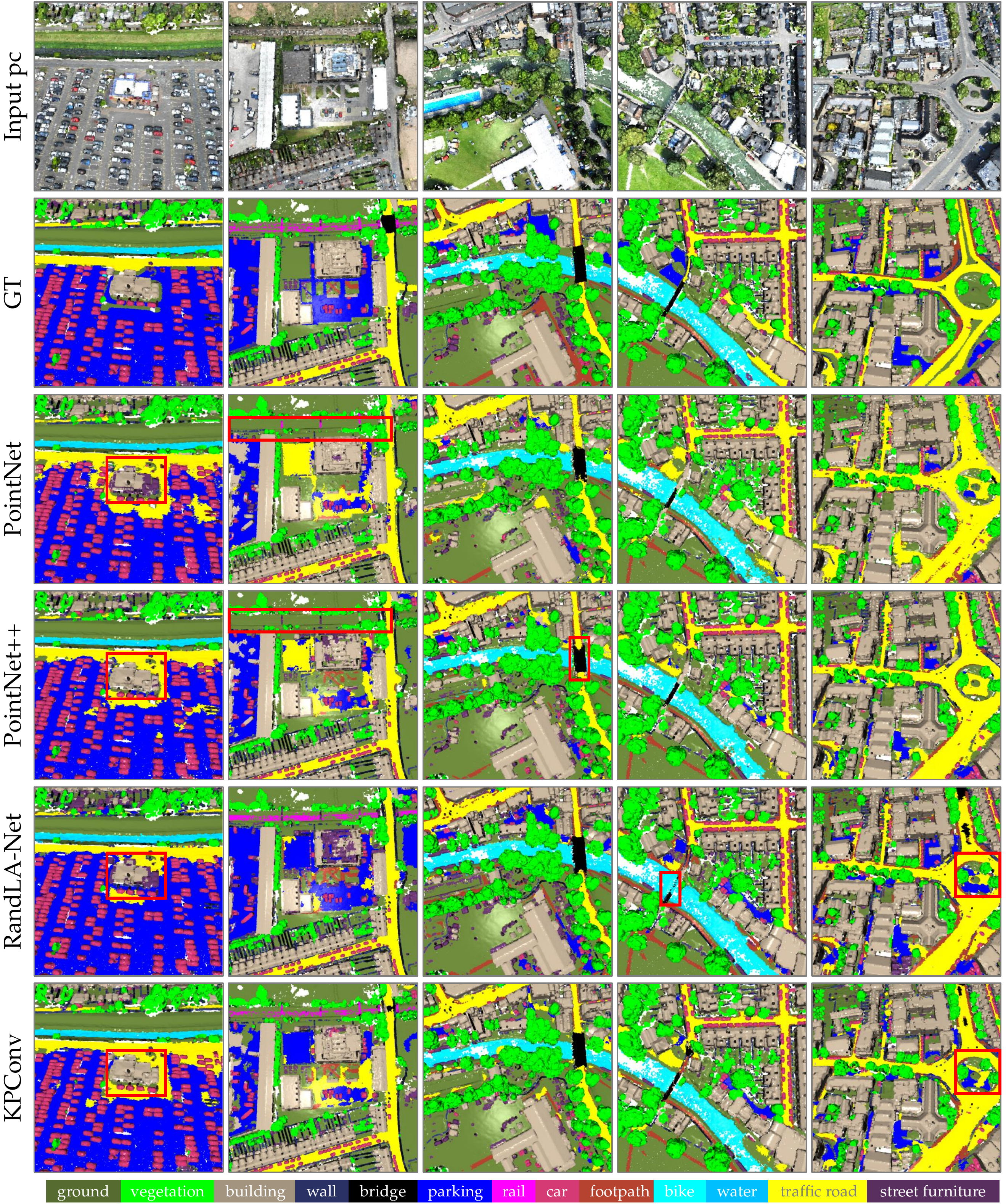}
\caption{Qualitative results of PointNet \cite{qi2017pointnet}, PointNet++ \cite{qi2017pointnet++}, RandLA-Net \cite{hu2019randla} and KPConv \cite{thomas2019kpconv}  on the test set of \nicknameData{} dataset. The black dashed box highlights the inconsistency predictions with the ground-truth label.}
\label{fig:Fig_supp_results}
\end{figure*}

\subsection{Geometry vs. Appearance}
\label{color}
Different from the point clouds acquired by the airbone LiDAR sensors \cite{varney2020dales,NPM3D,behley2019semantickitti,zolanvari2019dublincity} , the point clouds in our \nicknameData{} are colorized with fine-grained point-wise RGB information. Intuitively, the additional color features can provide informative appearance, further enable existing neural architectures to distinguish between heterogeneous semantic categories with similar geometrical structure (\textit{e.g.}, grass on the ground). However, the additional color information may also introduce distractors, which in turn deteriorate the final performance.

Existing techniques usually integrated the RGB color as additional channels of the input feature map feed into the network. However, the recent ShellNet \cite{zhang2019shellnet} learn the semantics from the pure spatial coordinates, but also achieves surprisingly good results. Overall, it remains an open question whether, and how, the color information impacts the final segmentation performance. To this end, we further conduct  comparative experiments to verify the impact of the color information to the final segmentation performance. In particular, five baselines including  PointNet/PointNet++ \cite{qi2017pointnet,qi2017pointnet++}, SPGraph \cite{landrieu2018large}, KPConv \cite{thomas2019kpconv}, and RandLA-Net \cite{hu2019randla} are selected for 10 groups of experiments. Each baseline is trained with the pure geometrical information (\textit{i.e.}, 3D coordinates) or both 3D coordinates and RGB appearance, separately.

\textbf{Analysis.} 
We report the quantitative results achieved by the selected five baselines with/without the usage of color in the input point clouds. We can see that:

\begin{itemize}[leftmargin=*]
\item All of PointNet/PointNet++, KPConv, and RandLA-Net achieve significant performance improvement when the color features are utilized, compared with the use geometrical coordinates alone. It is noted that categories with significant performance improvements include \textit{bridge}, \textit{footpath}, and \textit{water}, since these categories are geometrically indistinguishable.

\item We also noticed that the performance improvement of SPGraph is relatively marginal (only 2\%) compared with other baseline approaches. This is likely due to the homogenous geometrical partition used in its framework, which purely relies on the geometrical structure but ignores the informative color.
\end{itemize}

Apart from the quantitative results, we also explicitly visualize the qualitative results achieved by these baselines in Figure \ref{fig:Fig_supp_results}. To summarize, our experiments highlight the importance of color information to the fine-grained understanding of urban-scale point clouds, reflecting the advantages of our SensatUrban dataset over other existing aerial point clouds datasets collected by LiDAR, such as DALES \cite{varney2020dales}, NPM3D \cite{NPM3D}, and DublinCity \cite{zolanvari2019dublincity}. In particular, the color information is particularly important for distinguishing heterogeneous categories with consistent geometric structures (e.g., grass on the road), enabling a higher level of semantic understanding. This also provides insights for future aerial mapping campaigns, where color information and even other spectral bands may be useful for the semantic understanding.

\begin{table*}[thb]
\centering
\caption{Cross-city generalization performance of selected baselines on our \nicknameData{} dataset. All baselines are trained on the \textit{training split of Birmingham}. The top five records show the testing results on the \textit{testing split of Birmingham}, while the bottom five rows show the scores on the \textit{testing split of Cambridge (cross-city)}.\label{tab:cross-city}}
\resizebox{\textwidth}{!}{%
\begin{tabular}{lcccccccccccccccc}
\Xhline{2.0\arrayrulewidth}
 & \rotatebox{90}{OA(\%)} &  \rotatebox{90}{mAcc(\%)} & \rotatebox{90}{\textbf{mIoU(\%)}} & \rotatebox{90}{ground} & \rotatebox{90}{veg.} & \rotatebox{90}{building} & \rotatebox{90}{wall} & \rotatebox{90}{bridge} & \rotatebox{90}{parking} & \rotatebox{90}{rail} & \rotatebox{90}{traffic.} & \rotatebox{90}{street.} & \rotatebox{90}{car} & \rotatebox{90}{footpath} & \rotatebox{90}{bike} & \rotatebox{90}{water} \\
\Xhline{1.25\arrayrulewidth} 
PointNet \cite{qi2017pointnet} & 87.33 & 54.76 & 48.73 & 80.91 & 94.58 & 87.40 & 33.69 & 0.51 & 66.23 & 16.98 & 49.55 & 36.08 & 74.59 & 1.49 & 0.00 & 91.51\\
PointNet++ \cite{qi2017pointnet++} & 89.85 & 64.24 & 57.39 & 84.34 & 97.11 & 89.74 & 61.56 & \textbf{3.78} & 68.08 & 41.95 & 54.43 & 51.54 & 84.73 & 14.43 & 0.00 & \textbf{94.34}\\
SPGraph \cite{landrieu2018large} & 80.13 & 42.87 & 36.95 & 65.75 & 93.33 & 87.24 & 41.28 & 0.00 & 42.69 & 20.94 & 2.28 & 32.05 & 64.06 & 0.00 & 0.00 & 30.76\\
KPConv \cite{thomas2019kpconv} & \textbf{91.44} & 68.41 & \textbf{61.65} & \textbf{86.00} & \textbf{97.66} & \textbf{92.90} & \textbf{75.07} & 0.91 & \textbf{69.74} & \textbf{55.50} & \textbf{57.94} & \textbf{60.73} & \textbf{89.48} & 21.44 & 0.00 & 94.13\\
RandLA-Net \cite{hu2019randla} & 90.77 & \textbf{72.11} & 59.72 & 85.14 & 96.89 & 90.77 & 59.45 & 1.52 & 75.83 & 48.88 & 62.58 & 48.65 & 86.31 & \textbf{28.82} & 0.00 & 91.51\\
\midrule[0.5pt] 
PointNet \cite{qi2017pointnet} & 86.06 & 38.56 & 29.70 & 74.94 & 94.57 & 85.38 & 8.62 & 13.42 & 16.47 & 0.00 & 38.64 & 14.27 & 36.96 & 0.09 & 0.00 & 2.75\\
PointNet++ \cite{qi2017pointnet++} & 89.46 & 44.64 & 36.93 & 77.68 & 97.28 & 91.95 & 54.59 & 0.52 & 15.84 & 0.00 & 42.08 & 29.00 & 67.71 & 0.24 & 0.00 & 3.16\\
SPGraph \cite{landrieu2018large} & 82.02 & 24.83 & 20.70 & 61.72 & 88.26 & 78.27 & 8.29 & 0.00 & 0.00 & 0.00 & 0.64 & 1.87 & 30.00 & 0.00 & 0.00 & 0.00\\
KPConv \cite{thomas2019kpconv} & \textbf{90.62} & 48.71 & \textbf{40.51} & \textbf{78.88} & \textbf{98.33} & \textbf{94.24} & \textbf{76.20} & 0.01 & 14.70 & 0.00 & 41.77 & \textbf{39.32} & \textbf{74.22} & 0.39 & 0.00 & \textbf{8.61}\\
RandLA-Net \cite{hu2019randla} & 88.92 & \textbf{51.57} & 40.29 & 78.46 & 97.12 & 89.93 & 46.77 & \textbf{28.76} & \textbf{20.03} & 0.00 & \textbf{46.98} & 18.70 & 65.99 & \textbf{24.91} & 0.00 & 6.15\\
\Xhline{2.0\arrayrulewidth}
\end{tabular}}
\end{table*}

\begin{table*}[thb]
\centering
\caption{Cross-city generalization performance of selected baselines on our dataset. All baselines are trained on the \textit{training split of Cambridge}. The top five records show the testing results on the \textit{testing split of Cambridge}, while the bottom five rows show the scores on the \textit{testing split of Birmingham (cross-city)}. }\label{tab:cam2birm}
\resizebox{\textwidth}{!}{%
\begin{tabular}{lcccccccccccccccc}
\Xhline{2.0\arrayrulewidth}
 & \rotatebox{90}{OA(\%)} &  \rotatebox{90}{mAcc(\%)} & \rotatebox{90}{\textbf{mIoU(\%)}} & \rotatebox{90}{ground} & \rotatebox{90}{veg.} & \rotatebox{90}{building} & \rotatebox{90}{wall} & \rotatebox{90}{bridge} & \rotatebox{90}{parking} & \rotatebox{90}{rail} & \rotatebox{90}{traffic.} & \rotatebox{90}{street.} & \rotatebox{90}{car} & \rotatebox{90}{footpath} & \rotatebox{90}{bike} & \rotatebox{90}{water} \\
\Xhline{1.25\arrayrulewidth} 
PointNet \cite{qi2017pointnet} & 91.16 & 50.02 & 43.61 & 82.83 & 97.89 & 90.93 & 9.54 & 38.34 & 12.07 & 0.00 & 50.60 & 21.42 & 60.74 & 26.18 & 0.00 & 76.42 \\ 
PointNet++ \cite{qi2017pointnet++} & 93.62 & 62.64 & 55.60 & 85.50 & 98.93 & 95.35 & 63.73 & 59.19 & 24.00 & 0.00 & 59.13 & 40.50 & 79.30 & 38.01 & 0.00 & 79.11 \\ 
SPG \cite{landrieu2018large} & 84.13 & 36.36 & 31.55 & 68.96 & 92.14 & 84.61 & 15.37 & 13.84 & 4.46 & 0.00 & 31.83 & 21.02 & 22.04 & 0.47 & 0.00 & 23.83 \\ 
KPConv \cite{thomas2019kpconv} & \textbf{94.89} & \textbf{69.65} & \textbf{62.36} & \textbf{87.91} & \textbf{99.22} & \textbf{97.00} & \textbf{80.09} & \textbf{77.31} & \textbf{36.65} & 0.00 & \textbf{65.62} & \textbf{54.70} & \textbf{84.59} & \textbf{43.25} & \textbf{0.00} & \textbf{84.34} \\
RandLA-Net \cite{hu2019randla} & 91.45 & 69.55 & 53.21 & 81.39 & 98.49 & 93.43 & 56.40 & 49.40 & 35.80 & 0.00 & 60.75 & 31.29 & 81.11 & 37.20 & 0.00 & 66.55 \\ 
\midrule[0.5pt] 
PointNet \cite{qi2017pointnet} & 71.99 & 43.90 & 35.01 & 64.55 & 93.76 & 72.71 & 5.30 & 17.55 & 8.08 & 0.00 & 26.60 & 8.87 & 65.35 & 11.34 & 0.00 & \textbf{80.99} \\ 
PointNet++ \cite{qi2017pointnet++} & 78.47 & 52.32 & 41.29 & 70.52 & 96.07 & 70.13 & 44.89 & 6.60 & 34.67 & 0.00 & 33.39 & 27.42 & 73.79 & 16.78 & 0.00 & 62.48 \\ 
SPG \cite{landrieu2018large} & 71.27 & 28.42 & 22.93 & 57.32 & 84.28 & 76.51 & 12.40 & 8.95 & 0.00 & 0.00 & 20.41 & 10.52 & 14.08 & 0.00 & 0.00 & 13.65 \\ 
KPConv \cite{thomas2019kpconv} & \textbf{86.03} & \textbf{61.76} & \textbf{50.67} & \textbf{78.46} & \textbf{97.20} & 81.72 & \textbf{55.76} & \textbf{40.08} & \textbf{64.05} & 0.00 & \textbf{44.99} & \textbf{38.91} & \textbf{80.01} & \textbf{17.79} & \textbf{0.00} & 59.76 \\
RandLA-Net \cite{hu2019randla} & 75.52 & 59.27 & 40.08 & 63.42 & 93.12 & 73.54 & 44.09 & 5.35 & 45.54 & 0.00 & 31.68 & 25.66 & 74.47 & 8.93 & 0.00 & 55.25 \\ 
\Xhline{2.0\arrayrulewidth}
\end{tabular}%
}
\end{table*}

\subsection{The Impact of Skewed Class Distribution}
Although data preparations and the usage of color information have been considered, it is still noted that the performance of different semantic categories varies greatly. For example, all baseline methods can achieve excellent segmentation performance on \textit{vegetation}, with IoU scores up to 99\%, while completely failed in detecting rare patterns such as \textit{bikes}. Fundamentally, this is because of the extremely imbalanced distribution of our dataset. As illustrated in Figure \ref{fig:statistics}, the \nicknameData{} dataset is dominated by categories such as \textit{ground}/\textit{vegetation}/\textit{building}, which are commonly appeared in urban areas of modern cities. However, categories such as \textit{rail}/\textit{bike}, despite being highly important for infrastructure-oriented applications, occurring much less frequently than the prevalent categories. As consequence, the selected baselines show a biased tendency towards the prevalent categories during inference, due to the scarce occurrence of the under-represented categories.

It remains an open question to learn from training data with skewed distributions. In this paper, we attempt to alleviate this problem from the perspective of the loss function. In particular, advanced loss functions are utilized to adaptively re-weight the contributions of each point that belongs to different categories, eventually guiding the network to achieve a more balanced performance across different categories. Specifically, by taking PointNet and RandLA-Net as baselines, we replace the original vanilla cross-entropy loss with four off-the-shelf loss functions, including weighted cross-entropy with inverse frequency \cite{salsanext}, or with inverse square root (sqrt) frequency \cite{rosu2019latticenet}, \textit{Lov\'{a}sz}-softmax loss \cite{berman2018lovasz}, and focal loss \cite{focal_loss}.

\textbf{Analysis.} We report the detailed segmentation performance of two baselines achieved with five different loss functions. We can see the performance of all baselines has been improved when advanced loss functions are adopted. This clearly demonstrated that the sophisticated loss functions are indeed effective to alleviate the problem of imbalanced class distribution. Notably, we also noticed that the mIoU score of RandLA-Net has been improved by 5\% when using the weighted cross-entropy loss. In particular, the score in the most under-represented category bike has been improved by more than 20\%. Although the performance on minority categories is still far from satisfactory, the improvement is considerably encouraged and we suggest that more research could be conducted on this challenge, so as to fully tackle this research problem.

\subsection{Cross-City Generalization}
One of the main challenges of existing deep neural architectures is how to enhance the generalization capability to unseen scenarios, especially out-of-distribution data, since neural networks are usually data hungry and tend to overfit the training data. Motivated by this, we further explore the generalization performance of existing representative baselines on our SensatUrban dataset, since our dataset is composed of data collected from different cities, hence naturally suitable for evaluation the generalization abilities. In particular, five baseline approaches are included in our generalization experiments, that is: PointNet/PointNet++ \cite{qi2017pointnet,qi2017pointnet++}, SPGraph \cite{landrieu2018large}, KPConv \cite{thomas2019kpconv}, and RandLA-Net \cite{hu2019randla}. The detailed four groups of experimental schemes are described as follows:

\begin{itemize}[leftmargin=*]
\item Group 1: Train Birmingham/Test Birmingham. All of the five baseline approaches are only trained and tested on the training split and test split of Birmingham, respectively.
\item Group 2: Generalize from Birmingham to Cambridge: All of the five baselines are trained on the training split of Birmingham, and tested on the testing split of Cambridge. 
\item Group 3: Train Cambridge/Test Cambridge: Analogous to group 1, all of the 5 baselines are only trained on the training split of Cambridge, and then tested on the testing split of the same region.
\item Group 4: Generalize from Cambridge to Birmingham: the above well-trained 5 baseline models in group 3 are directly tested on the testing split of Birmingham. 
\end{itemize}

\textbf{Analysis.} We report the detailed results achieved in the first two groups of experiments in Table \ref{tab:cross-city}, and the last two groups of experiments in Table \ref{tab:cam2birm}. It can be seen that the performance of all baselines shows a significant decrease (approximate 20\% drop on average in mIoU scores) when generalizing the trained model to unseen urban areas in other cities, despite the data collected from different cities are actually in the same domain (\textit{i.e.,} captured using the same sensor). This demonstrates the limited generalization capacity of selected baseline approaches. Interestingly, we also noticed that the performance of dominant semantic categories such as \textit{vegetation} and \textit{building} are not severely affected, while the under-presented categories including \textit{rail} and \textit{water} show visible performance degradation. This shows the baseline approaches are actually overfitted to the prevalent categories, while they failed to learn generalized and meaningful representation for minority categories. Overall, generalizing the trained deep segmentation model to unseen data, especially point clouds with different distributions, remains an open question in this area. Therefore, we hope our SensatUrban dataset could highlight the limited generalization capacity of existing deep neural architectures, and inspire more research to be conducted on this challenging problem.

\begin{table}[thb]
\centering
\caption{The class mapping from DALES and \nicknameData{} dataset to the final unified semantic categories.}
\resizebox{0.45\textwidth}{!}{%
\begin{tabular}{ccc}
\Xhline{2.0\arrayrulewidth}
Classes of DALES & Mapped class & Classes of SensatUrban \\ 
\Xhline{1.25\arrayrulewidth} 
Ground & Ground & \begin{tabular}[c]{@{}c@{}}Ground, Bridge, Parking, \\ Rail, Traffic Road, Footpath\end{tabular} \\
Vegetation & Vegetation & Vegetation \\
Cars, Trucks & Cars & Cars \\
Power lines, Poles & Street furniture & Street furniture \\
Fences & Fences & Walls \\
Buildings & Buildings & Buildings \\
Unclassified & Unclassified (Ignored) & Bikes, Waters \\ 
\Xhline{2.0\arrayrulewidth}
\end{tabular}%
}
\label{tab:class-remapping}
\end{table}

\begin{table}[thb]
\centering
\caption{Statistics of the DALES dataset and SensatUrban dataset after class mapping.}
\resizebox{0.45\textwidth}{!}{%
\begin{tabular}{ccccc}
\Xhline{2.0\arrayrulewidth}
\multirow{2}{*}{Mapped classes} & \multicolumn{2}{c}{DALES} & \multicolumn{2}{c}{SensatUrban} \\ \cline{2-5} 
 & Training & Test & Training & Test \\ 
\Xhline{1.25\arrayrulewidth} 
Ground & 178,021,561 & 68,871,897 & 667,443,997 & 188,848,584 \\
Vegetation & 120,818,120 & 41,464,228 & 544,284,286 & 158,512,452 \\
Cars & 3,332,171 & 1,224,696 & 37,557,130 & 10,918,466 \\
Street furniture & 1,076,810 & 323,136 & 26,669,467 & 6,656,919 \\
Fences & 1,512,927 & 624,069 & 20,606,217 & 5,682,668 \\
Buildings & 56,908,533 & 23,454,294 & 861,977,674 & 164,867,233 \\
Unclassified & 6,997,560 & 681,571 & 7,262,966 & 4,449,979 \\ 
\Xhline{2.0\arrayrulewidth}
\end{tabular}%
}
\label{tab:statistics_class_map}
\end{table}

\subsection{Cross-Dataset Generalization}
Another interesting question is whether the deep model trained on our \nicknameData{} dataset can be well generalized to other similar airborne point cloud datasets \cite{varney2020dales, zolanvari2019dublincity}, or vice versa. Intuitively, this task seems even more challenging than cross-city generalization, since the point clouds acquired from different cities in our dataset are inherently homogeneous. That is, the point cloud is reconstructed from sequential aerial images captured by the same camera, with the identical data process pipeline. However, point clouds in different datasets are likely to be collected by distinct acquisition sensors (\ie airborne LiDAR vs. photogrammetry camera) and generated by using different mapping techniques. Moreover, the data distribution, point density, geographic regions, scene contents and annotation practices may vary greatly. Albeit interesting, there are few relevant studies in the field of 3D point cloud semantic understanding.

In this paper, we move a step forward to explore how the domain shift in different datasets affects the semantic learning of deep neural networks. Specifically, we select the recent aerial LiDAR point clouds dataset DALES \cite{varney2020dales}, along with the proposed photogrammetry \nicknameData{} dataset, to evaluate the cross dataset generalization capacity of existing segmentation algorithms. Note that, due to the inconsistent taxonomies and annotation practice, the semantic categories (\ie 8 valid categories in DALES vs. 13 valid categories in \nicknameData{}) and definitions are different in these two datasets. Therefore, we first reconcile the taxonomies and map the semantic categories into the newly defined 6 consistent semantic categories, so as to properly evaluate the generalization performance across datasets. The detailed class mapping from each dataset to the unified taxonomy is shown in Table \ref{tab:class-remapping}. The statistics (\ie the number of points in the training and test subset) after class mapping is reported in Table \ref{tab:statistics_class_map}. It can be seen that the class distribution of the two datasets exhibits visible differences. Here, we select the representative PointNet and RandLA-Net as the baselines, for the evaluation of intra-dataset generalization, and cross-dataset generalization performance. Note that, the baseline approaches are trained with  the usage of 3D spatial coordinates only, since the color information is not available in LiDAR point clouds provided by the DALES dataset.

\begin{table*}[thb]
\centering
\caption{Quantitative cross-dataset generalization results were achieved by the selected baseline approaches on the proposed SensatUrban dataset and the DALES dataset.}
\resizebox{1.0\textwidth}{!}{%
\begin{tabular}{llcccccccc}
\Xhline{2.0\arrayrulewidth}
Methods & Settings & OA(\%) & \textbf{mIoU(\%)} & \textit{Ground} & \textit{Vegetation} & \textit{Cars} & \textit{Street furniture} & \textit{Fences} & \textit{Buildings} \\ 
\Xhline{1.25\arrayrulewidth} 
\multirow{4}{*}{PointNet \cite{qi2017pointnet}} & DALES $\rightarrow$ DALES & 94.10 & 59.72 & 94.68 & 86.69 & 16.48 & 73.62 & 0.00 & 86.87 \\
 & DALES $\rightarrow$ SensatUrban & 74.25 & 30.75 & 89.44 & 55.69 & 0.02 & 0.03 & 0.00 & 39.32 \\
 & SensatUrban $\rightarrow$ SensatUrban & 92.46 & 56.27 & 92.90 & 92.14 & 52.69 & 0.33 & 14.25 & 85.33 \\
 & SensatUrban $\rightarrow$ DALES & 87.45 & 41.98 & 92.64 & 72.15 & 2.77 & 11.79 & 8.31 & 64.23 \\ 
\Xhline{2.0\arrayrulewidth}
\multirow{4}{*}{RandLA-Net \cite{hu2019randla}} & DALES $\rightarrow$ DALES & 96.98 & 84.31 & 96.99 & 92.71 & 80.54 & 89.08 & 50.09 & 96.47 \\
 & DALES $\rightarrow$ SensatUrban & 83.69 & 40.69 & 93.02 & 64.03 & 0.25 & 0.23 & 16.63 & 69.96 \\
 & SensatUrban $\rightarrow$ SensatUrban & 96.55 & 79.47 & 96.87 & 98.28 & 80.44 & 45.18 & 60.92 & 95.16 \\
 & SensatUrban$\rightarrow$DALES & 84.25 & 43.57 & 92.63 & 66.26 & 27.33 & 2.27 & 8.89 & 64.07 \\ 
\Xhline{2.0\arrayrulewidth}
\end{tabular}%
}
\label{tab:croos-dataset}
\end{table*}

\textbf{Analysis.} We can see that:  1) RandLA-Net has achieved superior performance in the intra-dataset evaluation, since the overall difficulty is reduced after the class mapping. 2) Although we considered the point density and the number of input points during data preprocessing, the cross-dataset generalization performance of both two baseline approaches is still significantly lower than the intra-dataset evaluation (30\%+), demonstrating that domain shifts in different datasets play a key role in preventing model generalization. Future studies are encouraged to further reduce the domain gap between different point cloud datasets, especially in light of the different configurations of existing LiDAR point clouds.

\subsection{Semantic Learning with Fewer Labels}
Deep learning-based methods are hungry for massive training data \cite{mprm}. For fully-supervised segmentation pipelines such as  \cite{hu2019randla, thomas2019kpconv, qi2017pointnet, qi2017pointnet++}, a large amount of fine-grained per-point annotations are usually required. However, it is extremely time-consuming and labor-intensive to manually annotate an urban-scale point cloud dataset with thousands of millions of points in practice. To this end, we further investigate the possibility of semantic learning with limited annotations on our \nicknameData{} dataset.

\begin{table*}[thb]
\centering
\caption{Quantitative results achieved by PointNet \cite{qi2017pointnet}, PointNet++ \cite{qi2017pointnet++}, and RandLA-Net \cite{hu2019randla} with different settings (varying number of semantic annotations). $^\dagger$means randomly a tiny fraction points of each semantic category, $^\ddagger$means randomly a tiny fraction points of the whole points.}
\label{tab:label-efficiency}
\resizebox{\textwidth}{!}{%
\begin{tabular}{clcccccccccccccccc}
\Xhline{2.0\arrayrulewidth}
Settings & Methods & \rotatebox{90}{OA(\%)} &  \rotatebox{90}{mAcc(\%)} & \rotatebox{90}{\textbf{mIoU(\%)}} & \rotatebox{90}{ground} & \rotatebox{90}{veg.} & \rotatebox{90}{building} & \rotatebox{90}{wall} & \rotatebox{90}{bridge} & \rotatebox{90}{parking} & \rotatebox{90}{rail} & \rotatebox{90}{traffic.} & \rotatebox{90}{street.} & \rotatebox{90}{car} & \rotatebox{90}{footpath} & \rotatebox{90}{bike} & \rotatebox{90}{water} \\
\Xhline{1.25\arrayrulewidth} 
\multirow{3}{*}{1 pt} & PointNet & 42.30 & 13.12 & 7.90 & 13.79 & 45.84 & 36.53 & 0.36 & 0.00 & 2.49 & 0.00 & 0.00 & 1.58 & 1.98 & 0.09 & 0.00 & 0.00 \\
 & PointNet++ & 29.52 & 20.82 & 8.74 & 21.48 & 24.52 & 24.80 & 1.81 & 0.00 & 8.99 & 0.00 & 15.56 & 3.52 & 5.91 & 7.02 & 0.02 & 0.03 \\
 & RandLA-Net & 1.97 & 14.07 & 1.93 & 0.00 & 0.00 & 0.00 & 1.31 & 0.59 & 7.92 & 0.30 & 0.02 & 1.17 & 1.00 & 7.84 & 0.01 & 4.94 \\
\Xhline{1.25\arrayrulewidth} 
\multirow{3}{*}{\begin{tabular}[c]{@{}c@{}}1\% annotation$^\dagger$\\ per category\end{tabular}} & PointNet & 89.58 & 52.87 & 45.95 & 82.92 & 97.40 & 90.07 & 16.12 & 28.83 & 33.82 & 5.86 & 44.95 & 26.98 & 65.50 & 23.15 & 0.00 & 81.81 \\
 & PointNet++ & 91.70 & 59.21 & 52.05 & 85.42 & 98.67 & 93.96 & 59.48 & 24.50 & 37.21 & 0.00 & 49.45 & 39.23 & 78.91 & 30.04 & 0.00 & 79.76 \\
 & RandLA-Net & 91.18 & 71.76 & 55.80 & 82.07 & 98.26 & 94.12 & 54.04 & 55.28 & 55.25 & 0.01 & 57.55 & 40.47 & 83.51 & 35.47 & 0.00 & 69.40 \\
\Xhline{1.25\arrayrulewidth} 
\multirow{3}{*}{1\% annotation$^\ddagger$} & PointNet & 89.11 & 50.25 & 43.46 & 82.33 & 97.13 & 89.20 & 11.12 & 15.92 & 34.38 & 0.00 & 42.33 & 23.63 & 63.73 & 26.33 & 0.00 & 78.94 \\
 & PointNet++ & 91.75 & 58.31 & 51.49 & 83.90 & 98.58 & 94.16 & 60.96 & 13.74 & 41.71 & 0.00 & 50.74 & 40.64 & 78.71 & 24.80 & 0.00 & 81.39 \\
 & RandLA-Net & 90.53 & 69.92 & 54.21 & 81.03 & 98.26 & 93.32 & 53.24 & 53.61 & 50.36 & 0.00 & 54.31 & 37.87 & 82.77 & 34.47 & 0.00 & 65.50 \\
\Xhline{1.25\arrayrulewidth} 
\multirow{3}{*}{\begin{tabular}[c]{@{}c@{}}10\% annotation$^\dagger$\\ per category\end{tabular}} & PointNet & 89.67 & 53.73 & 47.31 & 81.82 & 97.49 & 89.70 & 18.83 & 43.13 & 30.20 & 9.48 & 47.62 & 29.39 & 64.77 & 25.21 & 0.00 & 77.39 \\
 & PointNet++ & 92.60 & 63.24 & 56.67 & 86.08 & 98.73 & 94.65 & 64.68 & 49.87 & 43.77 & 0.00 & 53.70 & 45.51 & 82.29 & 36.20 & 0.00 & 81.27 \\
 & RandLA-Net & 91.59 & 72.92 & 57.44 & 83.07 & 98.18 & 93.48 & 51.25 & \textbf{54.77} & \textbf{58.32} & 12.72 & \textbf{62.06} & 38.81 & \textbf{83.82} & \textbf{39.42} & 0.00 & 70.79 \\
\Xhline{1.25\arrayrulewidth} 
\multirow{3}{*}{10\% annotation$^\ddagger$} & PointNet & 88.64 & 51.51 & 43.78 & 81.80 & 97.26 & 88.51 & 16.97 & 33.58 & 16.67 & 1.27 & 43.30 & 24.44 & 65.37 & 24.14 & 0.00 & 75.77 \\
 & PointNet++ & 92.28 & 60.71 & 53.84 & 85.39 & 98.64 & 94.13 & 60.59 & 25.89 & 46.15 & 0.00 & 53.34 & 42.73 & 80.78 & 30.04 & 0.00 & 82.19 \\
 & RandLA-Net & 89.91 & 69.80 & 52.15 & 79.73 & 97.91 & 93.04 & 54.08 & 44.06 & 50.68 & 0.00 & 53.87 & 35.52 & 81.72 & 30.58 & 0.00 & 56.75 \\
\Xhline{1.25\arrayrulewidth} 
\multirow{3}{*}{\begin{tabular}[c]{@{}c@{}}Full \\ supervision\end{tabular}} & PointNet & 90.57 & 56.30 & 49.69 & 83.55 & 97.67 & 90.66 & 22.56 & 43.54 & 40.35 & 9.29 & 50.74 & 29.58 & 68.24 & 29.27 & 0.00 & 80.55 \\
 & PointNet++ & \textbf{93.10} & 64.96 & 58.13 & \textbf{86.38} & \textbf{98.76} & \textbf{94.72} & \textbf{65.91} & 50.41 & 50.53 & 0.00 & 58.40 & \textbf{46.95} & 82.31 & 38.40 & 0.00 & \textbf{82.88} \\
 & RandLA-Net & 91.24 & \textbf{74.68} & \textbf{58.14} & 82.23 & 98.39 & 92.69 & 56.62 & 49.00 & 54.19 & \textbf{25.10} & 60.98 & 38.69 & 83.42 & 38.74 & 0.00 & 75.80 \\
\Xhline{2.0\arrayrulewidth}
\end{tabular}%
}
\end{table*}

\begin{table*}[thb]
\centering
\caption{Quantitative results achieved by using OcCo \cite{wang2020pre}, Jigsaw \cite{jigsaw} and Random (Rand) initialization on the \nicknameData{} dataset, based on PointNet \cite{qi2017pointnet}, PCN \cite{yuan2018pcn} and DGCNN \cite{dgcnn} encoders. Note that, all the initialized weights are obtained by pre-training on the ModelNet40 \cite{3d_shapenets}, since these techniques are mainly designed for object-level classification and segmentation.}\label{tab:self-supervise}
\resizebox{\textwidth}{!}{%
\begin{tabular}{rcccccccccccccccc}
\Xhline{2.0\arrayrulewidth}
 & \rotatebox{90}{OA(\%)} &  \rotatebox{90}{mAcc(\%)} & \rotatebox{90}{\textbf{mIoU(\%)}} & \rotatebox{90}{ground} & \rotatebox{90}{veg.} & \rotatebox{90}{building} & \rotatebox{90}{wall} & \rotatebox{90}{bridge} & \rotatebox{90}{parking} & \rotatebox{90}{rail} & \rotatebox{90}{traffic.} & \rotatebox{90}{street.} & \rotatebox{90}{car} & \rotatebox{90}{footpath} & \rotatebox{90}{bike} & \rotatebox{90}{water} \\
\Xhline{1.25\arrayrulewidth} 
PointNet \cite{qi2017pointnet} & 86.29 & 53.33 & 45.10 & 80.05 & 93.98 & 87.05 & 23.05 & 19.52 & 41.80 & 3.38 & 43.47 & 24.20 & 63.43 & 26.86 & 0.00 & 79.53 \\ 
PointNet-Jigsaw \cite{jigsaw} & 87.38 & 56.97 & 47.90 & 83.36 & 94.72 & 88.48 & 22.87 & 30.19 & 47.43 & 15.62 & 44.49 & 22.91 & 64.14 & 30.33 & 0.00 & 77.88 \\ 
PointNet-OcCo \cite{wang2020pre} & 87.87 & 56.14 & 48.50 & 83.76 & 94.81 & 89.24 & 23.29 & 33.38 & 48.04 & 15.84 & 45.38 & 24.99 & 65.00 & 27.13 & 0.00 & 79.58 \\
\Xhline{1.25\arrayrulewidth} 
PCN \cite{yuan2018pcn} & 86.79 & 57.66 & 47.91 & 82.61 & 94.82 & 89.04 & 26.66 & 21.96 & 34.96 & 28.39 & 43.32 & 27.13 & 62.97 & 30.87 & 0.00 & 80.06 \\
PCN-Jigsaw \cite{jigsaw} & 87.32 & 57.01 & 48.44 & 83.20 & 94.79 & 89.25 & 25.89 & 19.69 & 40.90 & 28.52 & 43.46 & 24.78 & 63.08 & 31.74 & 0.00 & \textbf{84.42} \\
PCN-OcCo \cite{wang2020pre} & 86.90 & 58.15 & 48.54 & 81.64 & 94.37 & 88.21 & 25.43 & 31.54 & 39.39 & 22.02 & 45.47 & 27.60 & 65.33 & 32.07 & 0.00 & 77.99 \\
\Xhline{1.25\arrayrulewidth} 
DGCNN \cite{dgcnn}  & 87.54 & 60.27 & 51.96 & 83.12 & 95.43 & 89.58 & \textbf{31.84} & 35.49 & 45.11 & 38.57 & 45.66 & 32.97 & 64.88 & 30.48 & 0.00 & 82.34 \\
DGCNN-Jigsaw \cite{jigsaw} & 88.65 & 60.80 & 53.01 & \textbf{83.95} & \textbf{95.92} & 89.85 & 30.05 & \textbf{43.59} & 46.40 & 35.28 & 49.60 & 31.46 & 69.41 & \textbf{34.38} & 0.00 & 80.55\\
DGCNN-OcCo \cite{wang2020pre} & \textbf{88.67} & \textbf{61.35} & \textbf{53.31} & 83.64 & 95.75 & \textbf{89.96} & 29.22 & 41.47 & \textbf{46.89} & \textbf{40.64} & \textbf{49.72} & \textbf{33.57} & \textbf{70.11} & 32.35 & 0.00 & 79.74 \\ 
\Xhline{2.0\arrayrulewidth}
\end{tabular}}
\end{table*}

Inspired by the weak supervision setting proposed in \cite{xu2020weakly}, we have conducted six groups of experiments by training all baselines with different forms of semantic annotations (\ie weak supervisions) in our dataset.  For simplicity, we only adopt PointNet \cite{qi2017pointnet}, PointNet++ \cite{qi2017pointnet++} and RandLA-Net \cite{hu2019randla} as baseline networks in the following groups of experiment:
\begin{itemize}[leftmargin=*]
\item Only 1 point annotated per category in each point cloud.
\item Only 1\% points annotated per category in each point cloud.
\item Only 1\% points annotated in each point cloud (randomly).
\item 10\% points annotated per category in each point cloud.
\item 10\% points annotated in each point cloud (randomly).
\item 100\% (all) points annotated in each point cloud.
\end{itemize}

\textbf{Analysis.} 
Table \ref{tab:label-efficiency} shows the detailed quantitative results achieved by three baselines under different settings. It can be seen that:
\begin{itemize}[leftmargin=*]
\item Surprisingly, both PointNet, PointNet++, and RandLA-Net can achieve comparable performance with their fully-supervised counterpart, even when training with a small fraction of labeled points (\eg 1\% or 10\%). This implies that the existing per-point annotations may exist large information redundancy, it is possible to learn semantics with limited annotations.
\item The performance of all baselines in group 2 and group 4 are better than group 3 and group 5, demonstrating that randomly annotating a tiny fraction  (\eg 1\%, 10\%) of all points is inferior to randomly annotating a tiny fraction  of points in each semantic category. Although randomly annotating a tiny fraction of all points is more practical and feasible.
\item The segmentation performance of all baselines in group 1 (\ie with 1 point annotation) are far from satisfactory. Basically, the networks cannot converge, primarily because the supervision information is extremely insufficient. However, this is also one of the simplest and cheapest ways of annotation in practice. More studies should be conducted in this direction to further improve the performance.
\end{itemize}

Thanks to the availability of several large-scale point cloud datasets, the community can always assume that the amount of labeled training data is sufficient. However, we have demonstrated that comparable performance can also be achieved using the same architecture with limited semantic annotations, highlighting the great potential of weakly-supervised semantic segmentation frameworks. This motivates us to further investigate how to achieve better performance under limited annotation, and how to choose the best annotation strategy under fixed budgets.

\subsection{Self-supervised Pre-Training on 3D Point Clouds}
Pre-training a network on a rich source set in a self-supervised or unsupervised way has been demonstrated to be highly effective for high-level downstream tasks (\eg segmentation, detection) in several 2D vision tasks \cite{simclr, he2020momentum}. However, self-supervised pre-training on 3D point clouds is still in its infancy, only a handful of recent works \cite{jigsaw,wang2020pre,xie2020pointcontrast,zhang2021self, hou2020exploring, orientation} have started to explore self-supervised learning on unstructured 3D point clouds. In particular, all existing methods are still pre-trained on the object-level datasets (\eg ModelNet40 \cite{3d_shapenets}) or indoor scene-level datasets (\eg ScanNet \cite{scannet}). Considering the urban-scale property of \nicknameData{}, it is particularly suitable for verifying the effectiveness of the existing pretraining strategy on our dataset. 

To this end, we conducted three groups of experiments on our \nicknameData{} dataset to compare the performance of:
\begin{itemize}[leftmargin=*]
\item Pre-training with occlusion completion \cite{wang2020pre}.
\item Pre-training with context prediction (jigsaw) \cite{jigsaw}.
\item Training from scratch.
\end{itemize}
For simplicity, we faithfully follow the three baseline networks used in their original paper \cite{wang2020pre, jigsaw}, including PointNet \cite{qi2017pointnet}, PCN \cite{yuan2018pcn}, and DGCNN \cite{dgcnn}. The detailed experimental results are shown in Table \ref{tab:self-supervise}.

\textbf{Analysis.} From the results in Table \ref{tab:self-supervise} we can see that, although both baseline networks are purely pre-trained on the object-level point clouds in ModelNet40, the fine-tuning models can still achieve a certain performance improvement on our dataset. In particular, the performance of several minority categories, such as \textit{rail} and \textit{bridge}, has a significant performance improvement (up to nearly 10\%), primary because the pre-trained models are less prone to overfitting to the majority categories, compared to directly training from scratch. 

This further demonstrates the feasibility and potential of the self-supervised pre-training paradigm. However, the existing pre-training framework \cite{wang2020pre,jigsaw} are still limited to object-level point clouds, and it is non-trivial to be extended to large-scale point clouds. On the other hand, most of the existing pre-training schemes are based on auxiliary (pre-text) tasks. It is worth investigating how to leverage contrastive learning to achieve better performance on 3D point clouds. Finally, to further facilitate the research in this research area, we also release the unlabeled York point clouds, encouraging further research exploration on this part of the data.

\section{Discussion and Limitations}
Although the proposed SensatUrban dataset is currently the largest publicly available photogrammetric point cloud dataset, it is not without limitation. In general, instance annotation would be a meaningful addition to our dataset. However, due to the tremendous labeling effort of point-wise instance labels, we leave the integration of instance labels for future exploration. On the other hand, our dataset is reconstructed from the sequential aerial images captured by a single sensor (\ie camera), it would be interesting to further investigate the same-source data acquired by different sensors. For example, the data acquired from both a camera and a LiDAR system integrated on the same UAV platform \cite{H3D}.

\section{Summary and Outlook}
\label{sec:sum}

This paper introduces SensatUrban: an urban-scale photogrammetric point cloud dataset composed of 7.6 $km^2$ of urban areas in three UK cities, and nearly 3 billion richly annotated points (each with one of the 13 semantic categories). A comprehensive benchmark is also built based on this dataset and a number of selected representative baselines. In particular, extensive comparative experiments have revealed several challenges in generalizing existing semantic segmentation methods to urban-scale point clouds, including how to conduct data preparation, whether and how to utilize the color information, how to tackle with the extremely imbalanced class distribution, generalizing to unseen scenarios, and the potential of weakly/self-supervised learning techniques. Besides, extensive benchmarking results are conducted and in-depth analysis are also provided. In the future, we will further increase the scale and richness (\textit{i.e.}, instance annotation, corresponding 2D images) of our SensatUrban dataset. We hope that our SensatUrban dataset could be an immensely useful resource and a canonical benchmark to related research communities including 3D computer vision, earth vision and remote sensing, inspiring and supporting future advancing research in related areas.

\begin{acknowledgements}
   This work was supported by a China Scholarship Council (CSC) scholarship, Huawei UK AI Fellowship, and the UKRI Natural Environment Research Council (NERC) Flood-PREPARED project (NE/P017134/1). Bo Yang was partially supported by HK PolyU (P0034792) and Shenzhen Science and Technology Innovation Commission (JCYJ20210324120603011). The authors highly appreciate the Data Study Group (DSG) organised by the Alan Turing Institute and the GPU resources generously provided by the LAVA group led by Professor Yulan Guo in the Sun Yat-sen University, China. The authors would also like to thank the pre-training results provided by Hanchen Wang from the University of Cambridge.
\end{acknowledgements}

\clearpage
\bibliographystyle{spbasic}      
\bibliography{Mendeley}   

\begin{thebibliography}{100}
\providecommand{\natexlab}[1]{#1}
\providecommand{\url}[1]{{#1}}
\providecommand{\urlprefix}{URL }
\expandafter\ifx\csname urlstyle\endcsname\relax
  \providecommand{\doi}[1]{DOI~\discretionary{}{}{}#1}\else
  \providecommand{\doi}{DOI~\discretionary{}{}{}\begingroup
  \urlstyle{rm}\Url}\fi
\providecommand{\eprint}[2][]{\url{#2}}

\bibitem[{Aksoy et~al.(2019)Aksoy, Baci, and Cavdar}]{salsanet}
Aksoy EE, Baci S, Cavdar S (2019) Salsanet: Fast road and vehicle segmentation
  in {LiDAR} point clouds for autonomous driving. In: 2020 IEEE Intelligent
  Vehicles Symposium (IV), pp 926--932

\bibitem[{Armeni et~al.(2017)Armeni, Sax, Zamir, and Savarese}]{2D-3D-S}
Armeni I, Sax S, Zamir AR, Savarese S (2017) Joint 2{D}-3{D}-semantic data for
  indoor scene understanding. In: Proceedings of the IEEE/CVF International
  Conference on Computer Vision

\bibitem[{Behley et~al.(2019)Behley, Garbade, Milioto, Quenzel, Behnke,
  Stachniss, and Gall}]{behley2019semantickitti}
Behley J, Garbade M, Milioto A, Quenzel J, Behnke S, Stachniss C, Gall J (2019)
  {SemanticKITTI}: A dataset for semantic scene understanding of {LiDAR}
  sequences. In: Proceedings of the IEEE/CVF International Conference on
  Computer Vision, pp 9297--9307

\bibitem[{Berman et~al.(2018)Berman, Rannen~Triki, and
  Blaschko}]{berman2018lovasz}
Berman M, Rannen~Triki A, Blaschko MB (2018) The lov{\'a}sz-softmax loss: a
  tractable surrogate for the optimization of the intersection-over-union
  measure in neural networks. In: Proceedings of the IEEE/CVF International
  Conference on Computer Vision, pp 4413--4421

\bibitem[{Boulch(2019)}]{boulch2019generalizing}
Boulch A (2019) Generalizing discrete convolutions for unstructured point
  clouds. arXiv preprint arXiv:190402375

\bibitem[{Caesar et~al.(2020)Caesar, Bankiti, Lang, Vora, Liong, Xu, Krishnan,
  Pan, Baldan, and Beijbom}]{caesar2020nuscenes}
Caesar H, Bankiti V, Lang AH, Vora S, Liong VE, Xu Q, Krishnan A, Pan Y, Baldan
  G, Beijbom O (2020) nu{S}cenes: A multimodal dataset for autonomous driving.
  In: Proceedings of the IEEE/CVF International Conference on Computer Vision,
  pp 11621--11631

\bibitem[{Chang et~al.(2018)Chang, Dai, Funkhouser, Halber, Niebner, Savva,
  Song, Zeng, and Zhang}]{Matterport3D}
Chang A, Dai A, Funkhouser T, Halber M, Niebner M, Savva M, Song S, Zeng A,
  Zhang Y (2018) {Matterport3D}: Learning from {RGB-D} data in indoor
  environments. In: 7th IEEE International Conference on 3D Vision, 3DV 2017,
  pp 667--676

\bibitem[{Chang et~al.(2015)Chang, Funkhouser, Guibas, Hanrahan, Huang, Li,
  Savarese, Savva, Song, Su et~al.}]{chang2015shapenet}
Chang AX, Funkhouser T, Guibas L, Hanrahan P, Huang Q, Li Z, Savarese S, Savva
  M, Song S, Su H, et~al. (2015) Shape{N}et: An information-rich 3{D} model
  repository. arXiv preprint arXiv:151203012

\bibitem[{Chang et~al.(2019)Chang, Lambert, Sangkloy, Singh, Bak, Hartnett,
  Wang, Carr, Lucey, Ramanan et~al.}]{chang2019argoverse}
Chang MF, Lambert J, Sangkloy P, Singh J, Bak S, Hartnett A, Wang D, Carr P,
  Lucey S, Ramanan D, et~al. (2019) Argoverse: 3{D} tracking and forecasting
  with rich maps. In: Proceedings of the IEEE/CVF Conference on Computer Vision
  and Pattern Recognition, pp 8748--8757

\bibitem[{Chen et~al.(2020)Chen, Kornblith, Norouzi, and Hinton}]{simclr}
Chen T, Kornblith S, Norouzi M, Hinton G (2020) A simple framework for
  contrastive learning of visual representations. In: International Conference
  on Machine Learning, pp 1597--1607

\bibitem[{Cheng et~al.(2021)Cheng, Razani, Taghavi, Li, and Liu}]{2-S3Net}
Cheng R, Razani R, Taghavi E, Li E, Liu B (2021) 2-s3net: Attentive feature
  fusion with adaptive feature selection for sparse semantic segmentation
  network. In: Proceedings of the IEEE/CVF Conference on Computer Vision and
  Pattern Recognition, pp 12547--12556

\bibitem[{Choy et~al.(2019)Choy, Gwak, and Savarese}]{4dMinkpwski}
Choy C, Gwak J, Savarese S (2019) 4{D} spatio-temporal convnets: Minkowski
  convolutional neural networks. In: Proceedings of the IEEE/CVF International
  Conference on Computer Vision, pp 3075--3084

\bibitem[{Cortinhal et~al.(2020)Cortinhal, Tzelepis, and Aksoy}]{salsanext}
Cortinhal T, Tzelepis G, Aksoy EE (2020) Salsanext: Fast semantic segmentation
  of {LiDAR} point clouds for autonomous driving. arXiv preprint
  arXiv:200303653

\bibitem[{Dai et~al.(2017)Dai, Chang, Savva, Halber, Funkhouser, and
  Nie{\ss}ner}]{scannet}
Dai A, Chang AX, Savva M, Halber M, Funkhouser T, Nie{\ss}ner M (2017)
  Scan{N}et: Richly-annotated 3{D} reconstructions of indoor scenes. In:
  Proceedings of the IEEE/CVF International Conference on Computer Vision, pp
  5828--5839

\bibitem[{De~Deuge et~al.(2013)De~Deuge, Quadros, Hung, and
  Douillard}]{de2013unsupervised}
De~Deuge M, Quadros A, Hung C, Douillard B (2013) Unsupervised feature learning
  for classification of outdoor 3{D} scans. In: Australasian Conference on
  Robitics and Automation, vol~2, p~1

\bibitem[{Gaidon et~al.(2016)Gaidon, Wang, Cabon, and Vig}]{gaidon2016virtual}
Gaidon A, Wang Q, Cabon Y, Vig E (2016) Virtual worlds as proxy for
  multi-object tracking analysis. In: Proceedings of the IEEE/CVF International
  Conference on Computer Vision, pp 4340--4349

\bibitem[{Geiger et~al.(2012)Geiger, Lenz, and Urtasun}]{geiger2012we}
Geiger A, Lenz P, Urtasun R (2012) Are we ready for autonomous driving? the
  {KITTI} vision benchmark suite. In: 2012 IEEE Conference on Computer Vision
  and Pattern Recognition, pp 3354--3361

\bibitem[{Geiger et~al.(2013)Geiger, Lenz, Stiller, and
  Urtasun}]{geiger2013vision}
Geiger A, Lenz P, Stiller C, Urtasun R (2013) Vision meets robotics: The kitti
  dataset. The International Journal of Robotics Research 32(11):1231--1237

\bibitem[{Gerke and Kerle(2011)}]{damage_assessment}
Gerke M, Kerle N (2011) Automatic structural seismic damage assessment with
  airborne oblique pictometry{\copyright} imagery. Photogrammetric Engineering
  \& Remote Sensing 77(9):885--898

\bibitem[{Geyer et~al.(2020)Geyer, Kassahun, Mahmudi, Ricou, Durgesh, Chung,
  Hauswald, Pham, M{\"u}hlegg, Dorn et~al.}]{geyer2020a2d2}
Geyer J, Kassahun Y, Mahmudi M, Ricou X, Durgesh R, Chung AS, Hauswald L, Pham
  VH, M{\"u}hlegg M, Dorn S, et~al. (2020) A{2D}2: Audi autonomous driving
  dataset. arXiv preprint arXiv:200406320

\bibitem[{Graham et~al.(2018)Graham, Engelcke, and van~der Maaten}]{sparse}
Graham B, Engelcke M, van~der Maaten L (2018) 3{D} semantic segmentation with
  submanifold sparse convolutional networks. In: Proceedings of the IEEE/CVF
  International Conference on Computer Vision

\bibitem[{Guo et~al.(2020)Guo, Wang, Hu, Liu, Liu, and Bennamoun}]{guo2019deep}
Guo Y, Wang H, Hu Q, Liu H, Liu L, Bennamoun M (2020) Deep learning for 3{D}
  point clouds: A survey. IEEE TPAMI

\bibitem[{Hackel et~al.(2017)Hackel, Savinov, Ladicky, Wegner, Schindler, and
  Pollefeys}]{Semantic3D}
Hackel T, Savinov N, Ladicky L, Wegner JD, Schindler K, Pollefeys M (2017)
  Semantic3{D}.{N}et: A new large-scale point cloud classification benchmark.
  ISPRS Annals of Photogrammetry, Remote Sensing \& Spatial Information
  Sciences

\bibitem[{Han et~al.(2020)Han, Zheng, Xu, and Fang}]{han2020occuseg}
Han L, Zheng T, Xu L, Fang L (2020) Occuseg: Occupancy-aware 3d instance
  segmentation. In: Proceedings of the IEEE/CVF conference on computer vision
  and pattern recognition, pp 2940--2949

\bibitem[{Handa et~al.(2016)Handa, Patraucean, Badrinarayanan, Stent, and
  Cipolla}]{handa2015scenenet}
Handa A, Patraucean V, Badrinarayanan V, Stent S, Cipolla R (2016) Scene{N}et:
  understanding real world indoor scenes with synthetic data. In: Proceedings
  of the IEEE/CVF International Conference on Computer Vision

\bibitem[{He et~al.(2020)He, Fan, Wu, Xie, and Girshick}]{he2020momentum}
He K, Fan H, Wu Y, Xie S, Girshick R (2020) Momentum contrast for unsupervised
  visual representation learning. In: Proceedings of the IEEE/CVF Conference on
  Computer Vision and Pattern Recognition, pp 9729--9738

\bibitem[{Hou et~al.(2020)Hou, Graham, Nie{\ss}ner, and Xie}]{hou2020exploring}
Hou J, Graham B, Nie{\ss}ner M, Xie S (2020) Exploring data-efficient 3{D}
  scene understanding with contrastive scene contexts. arXiv preprint
  arXiv:201209165

\bibitem[{Hou et~al.(2014)Hou, Wang, Wang, Maynard, Cameron, Zhang, and
  Jiao}]{assest_management}
Hou L, Wang Y, Wang X, Maynard N, Cameron IT, Zhang S, Jiao Y (2014) Combining
  photogrammetry and augmented reality towards an integrated facility
  management system for the oil industry. Proceedings of the IEEE
  102(2):204--220

\bibitem[{Hu et~al.(2003)Hu, You, and Neumann}]{urban_planning}
Hu J, You S, Neumann U (2003) Approaches to large-scale urban modeling. IEEE
  Computer Graphics and Applications 23(6):62--69

\bibitem[{Hu et~al.(2020)Hu, Yang, Xie, Rosa, Guo, Wang, Trigoni, and
  Markham}]{hu2019randla}
Hu Q, Yang B, Xie L, Rosa S, Guo Y, Wang Z, Trigoni N, Markham A (2020)
  {RandLA-Net}: Efficient semantic segmentation of large-scale point clouds.
  In: Proceedings of the IEEE/CVF International Conference on Computer Vision

\bibitem[{Hu et~al.(2021)Hu, Yang, Khalid, Xiao, Trigoni, and
  Markham}]{hu2020towards}
Hu Q, Yang B, Khalid S, Xiao W, Trigoni N, Markham A (2021) Towards semantic
  segmentation of urban-scale 3d point clouds: A dataset, benchmarks and
  challenges. In: Proceedings of the IEEE/CVF Conference on Computer Vision and
  Pattern Recognition, pp 4977--4987

\bibitem[{Jiang et~al.(2020)Jiang, Zhao, Shi, Liu, Fu, and
  Jia}]{jiang2020pointgroup}
Jiang L, Zhao H, Shi S, Liu S, Fu CW, Jia J (2020) Pointgroup: Dual-set point
  grouping for 3d instance segmentation. In: Proceedings of the IEEE/CVF
  Conference on Computer Vision and Pattern Recognition, pp 4867--4876

\bibitem[{K{\"o}lle et~al.(2021)K{\"o}lle, Laupheimer, Schmohl, Haala,
  Rottensteiner, Wegner, and Ledoux}]{H3D}
K{\"o}lle M, Laupheimer D, Schmohl S, Haala N, Rottensteiner F, Wegner JD,
  Ledoux H (2021) H3d: Benchmark on semantic segmentation of high-resolution 3d
  point clouds and textured meshes from uav lidar and multi-view-stereo. arXiv
  preprint arXiv:210205346

\bibitem[{Landrieu and Simonovsky(2018)}]{landrieu2018large}
Landrieu L, Simonovsky M (2018) Large-scale point cloud semantic segmentation
  with superpoint graphs. In: Proceedings of the IEEE/CVF International
  Conference on Computer Vision, pp 4558--4567

\bibitem[{Lang et~al.(2019)Lang, Vora, Caesar, Zhou, Yang, and
  Beijbom}]{lang2019pointpillars}
Lang AH, Vora S, Caesar H, Zhou L, Yang J, Beijbom O (2019) Pointpillars: Fast
  encoders for object detection from point clouds. In: Proceedings of the
  IEEE/CVF Conference on Computer Vision and Pattern Recognition, pp
  12697--12705

\bibitem[{Le and Duan(2018)}]{le2018pointgrid}
Le T, Duan Y (2018) Point{G}rid: A deep network for 3{D} shape understanding.
  In: Proceedings of the IEEE/CVF International Conference on Computer Vision,
  pp 9204--9214

\bibitem[{Lei et~al.(2020)Lei, Akhtar, and Mian}]{SPH3D}
Lei H, Akhtar N, Mian A (2020) Spherical kernel for efficient graph convolution
  on 3{D} point clouds. IEEE Transactions on Pattern Analysis and Machine
  Intelligence

\bibitem[{Li et~al.(2020)Li, Li, Tong, Lim, Yuan, Wu, Tang, and
  Huang}]{li2020campus3d}
Li X, Li C, Tong Z, Lim A, Yuan J, Wu Y, Tang J, Huang R (2020) Campus3{D}: A
  photogrammetry point cloud benchmark for hierarchical understanding of
  outdoor scene. ACM MM

\bibitem[{Li et~al.(2018)Li, Bu, Sun, Wu, Di, and Chen}]{li2018pointcnn}
Li Y, Bu R, Sun M, Wu W, Di X, Chen B (2018) {PointCNN}: Convolution on
  {X}-transformed points. Advances in Neural Information Processing Systems

\bibitem[{Lin et~al.(2017)Lin, Goyal, Girshick, He, and
  Doll{\'a}r}]{focal_loss}
Lin TY, Goyal P, Girshick R, He K, Doll{\'a}r P (2017) Focal loss for dense
  object detection. Proceedings of the IEEE/CVF International Conference on
  Computer Vision

\bibitem[{Liu et~al.(2019)Liu, Tang, Lin, and Han}]{Point_voxel_cnn}
Liu Z, Tang H, Lin Y, Han S (2019) Point-voxel cnn for efficient 3{D} deep
  learning. Advances in Neural Information Processing Systems

\bibitem[{Lyu et~al.(2020)Lyu, Huang, and Zhang}]{lyu2020learning}
Lyu Y, Huang X, Zhang Z (2020) Learning to segment 3{D} point clouds in {2D}
  image space. In: Proceedings of the IEEE/CVF International Conference on
  Computer Vision

\bibitem[{McCormac et~al.(2016)McCormac, Handa, Leutenegger, and
  Davison}]{scenenetrgbd}
McCormac J, Handa A, Leutenegger S, Davison AJ (2016) {SceneNet RGB-D}: 5m
  photorealistic images of synthetic indoor trajectories with ground truth.
  arXiv preprint arXiv:161205079

\bibitem[{Meng et~al.(2019)Meng, Gao, Lai, and Manocha}]{vvnet}
Meng HY, Gao L, Lai YK, Manocha D (2019) {VV-Net}: Voxel vae net with group
  convolutions for point cloud segmentation. Proceedings of the IEEE/CVF
  International Conference on Computer Vision

\bibitem[{Milioto et~al.(2019)Milioto, Vizzo, Behley, and
  Stachniss}]{rangenet++}
Milioto A, Vizzo I, Behley J, Stachniss C (2019) Rangenet++: Fast and accurate
  {LiDAR} semantic segmentation. In: 2019 IEEE/RSJ International Conference on
  Intelligent Robots and Systems (IROS), pp 4213--4220

\bibitem[{Mo et~al.(2019)Mo, Zhu, Chang, Yi, Tripathi, Guibas, and
  Su}]{mo2019partnet}
Mo K, Zhu S, Chang AX, Yi L, Tripathi S, Guibas LJ, Su H (2019) {PartNet}: A
  large-scale benchmark for fine-grained and hierarchical part-level 3{D}
  object understanding. In: Proceedings of the IEEE/CVF Conference on Computer
  Vision and Pattern Recognition, pp 909--918

\bibitem[{Munoz et~al.(2009)Munoz, Bagnell, Vandapel, and Hebert}]{oakland}
Munoz D, Bagnell JA, Vandapel N, Hebert M (2009) Contextual classification with
  functional max-margin markov networks. In: Proceedings of the IEEE/CVF
  International Conference on Computer Vision

\bibitem[{{\"O}zdemir et~al.(2019){\"O}zdemir, Toschi, and Remondino}]{3DOM}
{\"O}zdemir E, Toschi I, Remondino F (2019) a multi-purpose benchmark for
  photogrammetric urban 3{D} reconstruction in a controlled environment. In:
  Evaluation and Benchmarking Sensors, Systems and Geospatial Data in
  Photogrammetry and Remote Sensing, vol~42, pp 53--60

\bibitem[{Pan et~al.(2020)Pan, Gao, Mei, Geng, Li, and
  Zhao}]{pan2020semanticposs}
Pan Y, Gao B, Mei J, Geng S, Li C, Zhao H (2020) Semanticposs: A point cloud
  dataset with large quantity of dynamic instances. arXiv preprint
  arXiv:200209147

\bibitem[{Poursaeed et~al.(2020)Poursaeed, Jiang, Qiao, Xu, and
  Kim}]{orientation}
Poursaeed O, Jiang T, Qiao Q, Xu N, Kim VG (2020) Self-supervised learning of
  point clouds via orientation estimation. arXiv preprint arXiv:200800305

\bibitem[{Qi et~al.(2017{\natexlab{a}})Qi, Su, Mo, and Guibas}]{qi2017pointnet}
Qi CR, Su H, Mo K, Guibas LJ (2017{\natexlab{a}}) {PointNet}: Deep learning on
  point sets for {3D} classification and segmentation. In: Proceedings of the
  IEEE/CVF International Conference on Computer Vision, pp 652--660

\bibitem[{Qi et~al.(2017{\natexlab{b}})Qi, Yi, Su, and
  Guibas}]{qi2017pointnet++}
Qi CR, Yi L, Su H, Guibas LJ (2017{\natexlab{b}}) Point{N}et++: Deep
  hierarchical feature learning on point sets in a metric space. Advances in
  Neural Information Processing Systems

\bibitem[{Qin et~al.(2021)Qin, Tan, Ma, Zhang, and Li}]{qin2021opengf}
Qin N, Tan W, Ma L, Zhang D, Li J (2021) Opengf: An ultra-large-scale ground
  filtering dataset built upon open als point clouds around the world. In:
  Proceedings of the IEEE/CVF Conference on Computer Vision and Pattern
  Recognition, pp 1082--1091

\bibitem[{Rao et~al.(2010)Rao, Le, Phoka, Quigley, Sudsang, and
  Ng}]{rao2010grasping}
Rao D, Le QV, Phoka T, Quigley M, Sudsang A, Ng AY (2010) Grasping novel
  objects with depth segmentation. In: 2010 IEEE/RSJ International Conference
  on Intelligent Robots and Systems, pp 2578--2585

\bibitem[{Ros et~al.(2016)Ros, Sellart, Materzynska, Vazquez, and
  Lopez}]{ros2016synthia}
Ros G, Sellart L, Materzynska J, Vazquez D, Lopez AM (2016) The {SYNTHIA}
  dataset: A large collection of synthetic images for semantic segmentation of
  urban scenes. In: Proceedings of the IEEE/CVF International Conference on
  Computer Vision, pp 3234--3243

\bibitem[{Rosu et~al.(2019)Rosu, Sch{\"u}tt, Quenzel, and
  Behnke}]{rosu2019latticenet}
Rosu RA, Sch{\"u}tt P, Quenzel J, Behnke S (2019) Lattice{N}et: Fast point
  cloud segmentation using permutohedral lattices. arXiv preprint
  arXiv:191205905

\bibitem[{Rottensteiner et~al.(2012)Rottensteiner, Sohn, Jung, Gerke, Baillard,
  Benitez, and Breitkopf}]{rottensteiner2012isprs}
Rottensteiner F, Sohn G, Jung J, Gerke M, Baillard C, Benitez S, Breitkopf U
  (2012) The {ISPRS} benchmark on urban object classification and 3{D} building
  reconstruction. ISPRS Annals of the Photogrammetry, Remote Sensing and
  Spatial Information Sciences I-3 (2012), Nr 1 1(1):293--298

\bibitem[{Roynard et~al.(2018)Roynard, Deschaud, and Goulette}]{NPM3D}
Roynard X, Deschaud JE, Goulette F (2018) {Paris-Lille-3D}: A large and
  high-quality ground-truth urban point cloud dataset for automatic
  segmentation and classification. The International Journal of Robotics
  Research 37(6):545--557

\bibitem[{Sauder and Sievers(2019)}]{jigsaw}
Sauder J, Sievers B (2019) Self-supervised deep learning on point clouds by
  reconstructing space. Advances in Neural Information Processing Systems pp
  12962--12972

\bibitem[{Serna et~al.(2014)Serna, Marcotegui, Goulette, and
  Deschaud}]{paris-rue-madame}
Serna A, Marcotegui B, Goulette F, Deschaud JE (2014) Paris-rue-madame
  database: a 3{D} mobile laser scanner dataset for benchmarking urban
  detection, segmentation and classification methods. In: 4th International
  Conference on Pattern Recognition, Applications and Methods ICPRAM 2014

\bibitem[{Silberman et~al.(2012)Silberman, Hoiem, Kohli, and Fergus}]{NYU3D}
Silberman N, Hoiem D, Kohli P, Fergus R (2012) Indoor segmentation and support
  inference from {RGBD} images. In: European conference on computer vision, pp
  746--760

\bibitem[{Song et~al.(2015)Song, Lichtenberg, and Xiao}]{sunrgbd}
Song S, Lichtenberg SP, Xiao J (2015) Sun {RGB-D}: A {RGB-D} scene
  understanding benchmark suite. In: Proceedings of the IEEE/CVF International
  Conference on Computer Vision, pp 567--576

\bibitem[{Sun et~al.(2020)Sun, Kretzschmar, Dotiwalla, Chouard, Patnaik, Tsui,
  Guo, Zhou, Chai, Caine et~al.}]{Waymo}
Sun P, Kretzschmar H, Dotiwalla X, Chouard A, Patnaik V, Tsui P, Guo J, Zhou Y,
  Chai Y, Caine B, et~al. (2020) Scalability in perception for autonomous
  driving: Waymo open dataset. In: Proceedings of the IEEE/CVF Conference on
  Computer Vision and Pattern Recognition, pp 2446--2454

\bibitem[{Tan et~al.(2020)Tan, Qin, Ma, Li, Du, Cai, Yang, and Li}]{Toronto3D}
Tan W, Qin N, Ma L, Li Y, Du J, Cai G, Yang K, Li J (2020) Toronto-3{D}: A
  large-scale mobile {LiDAR} dataset for semantic segmentation of urban
  roadways. In: Proceedings of the IEEE/CVF Conference on Computer Vision and
  Pattern Recognition Workshops, pp 202--203

\bibitem[{Tang et~al.(2020)Tang, Liu, Zhao, Lin, Lin, Wang, and Han}]{e3d}
Tang H, Liu Z, Zhao S, Lin Y, Lin J, Wang H, Han S (2020) Searching efficient
  3{D} architectures with sparse point-voxel convolution. In: European
  Conference on Computer Vision, pp 685--702

\bibitem[{Tatarchenko et~al.(2018)Tatarchenko, Park, Koltun, and
  Zhou}]{tangentconv}
Tatarchenko M, Park J, Koltun V, Zhou QY (2018) Tangent convolutions for dense
  prediction in 3{D}. In: Proceedings of the IEEE/CVF International Conference
  on Computer Vision, pp 3887--3896

\bibitem[{Tchapmi et~al.(2017)Tchapmi, Choy, Armeni, Gwak, and
  Savarese}]{tchapmi2017segcloud}
Tchapmi L, Choy C, Armeni I, Gwak J, Savarese S (2017) Segcloud: Semantic
  segmentation of {3D} point clouds. In: 2017 International Conference on 3D
  Vision (3DV), pp 537--547

\bibitem[{Thomas et~al.(2019)Thomas, Qi, Deschaud, Marcotegui, Goulette, and
  Guibas}]{thomas2019kpconv}
Thomas H, Qi CR, Deschaud JE, Marcotegui B, Goulette F, Guibas LJ (2019)
  {KPConv}: Flexible and deformable convolution for point clouds. Proceedings
  of the IEEE/CVF International Conference on Computer Vision pp 6411--6420

\bibitem[{Tong et~al.(2020)Tong, Li, Chen, Sun, Cao, and Xiang}]{tong2020cspc}
Tong G, Li Y, Chen D, Sun Q, Cao W, Xiang G (2020) {CSPC}-dataset: New {LiDAR}
  point cloud dataset and benchmark for large-scale scene semantic
  segmentation. IEEE Access

\bibitem[{Uy et~al.(2019)Uy, Pham, Hua, Nguyen, and Yeung}]{scanobjectnn}
Uy MA, Pham QH, Hua BS, Nguyen T, Yeung SK (2019) Revisiting point cloud
  classification: A new benchmark dataset and classification model on
  real-world data. In: Proceedings of the IEEE/CVF International Conference on
  Computer Vision, pp 1588--1597

\bibitem[{Valada et~al.(2017)Valada, Vertens, Dhall, and
  Burgard}]{valada2017adapnet}
Valada A, Vertens J, Dhall A, Burgard W (2017) Adapnet: Adaptive semantic
  segmentation in adverse environmental conditions. In: 2017 IEEE International
  Conference on Robotics and Automation (ICRA), pp 4644--4651

\bibitem[{Vallet et~al.(2015)Vallet, Br{\'e}dif, Serna, Marcotegui, and
  Paparoditis}]{IQmulus}
Vallet B, Br{\'e}dif M, Serna A, Marcotegui B, Paparoditis N (2015)
  {TerraMobilita/iQmulus} urban point cloud analysis benchmark. Computers \&
  Graphics

\bibitem[{Varney et~al.(2020)Varney, Asari, and Graehling}]{varney2020dales}
Varney N, Asari VK, Graehling Q (2020) {DALES}: A large-scale aerial {LiDAR}
  data set for semantic segmentation. In: Proceedings of the IEEE/CVF
  Conference on Computer Vision and Pattern Recognition Workshops, pp 186--187

\bibitem[{Wang et~al.(2020)Wang, Liu, Yue, Lasenby, and Kusner}]{wang2020pre}
Wang H, Liu Q, Yue X, Lasenby J, Kusner MJ (2020) Pre-training by completing
  point clouds. arXiv preprint arXiv:201001089

\bibitem[{Wang et~al.(2019{\natexlab{a}})Wang, Huang, Hou, Zhang, and
  Shan}]{GACNet}
Wang L, Huang Y, Hou Y, Zhang S, Shan J (2019{\natexlab{a}}) Graph attention
  convolution for point cloud semantic segmentation. In: Proceedings of the
  IEEE/CVF International Conference on Computer Vision

\bibitem[{Wang et~al.(2019{\natexlab{b}})Wang, Sun, Liu, Sarma, Bronstein, and
  Solomon}]{dgcnn}
Wang Y, Sun Y, Liu Z, Sarma SE, Bronstein MM, Solomon JM (2019{\natexlab{b}})
  Dynamic graph cnn for learning on point clouds. Acm Transactions On Graphics
  (TOG) 38(5):1--12

\bibitem[{Wei et~al.(2020)Wei, Lin, Yap, Hung, and Xie}]{mprm}
Wei J, Lin G, Yap KH, Hung TY, Xie L (2020) Multi-path region mining for weakly
  supervised 3{D} semantic segmentation on point clouds. In: Proceedings of the
  IEEE/CVF Conference on Computer Vision and Pattern Recognition, pp 4384--4393

\bibitem[{Westoby et~al.(2012)Westoby, Brasington, Glasser, Hambrey, and
  Reynolds}]{westoby2012structure}
Westoby MJ, Brasington J, Glasser NF, Hambrey MJ, Reynolds JM (2012)
  ‘structure-from-motion’photogrammetry: A low-cost, effective tool for
  geoscience applications. Geomorphology 179:300--314

\bibitem[{Wu et~al.(2018{\natexlab{a}})Wu, Wan, Yue, and
  Keutzer}]{wu2018squeezeseg}
Wu B, Wan A, Yue X, Keutzer K (2018{\natexlab{a}}) Squeeze{S}eg: Convolutional
  neural nets with recurrent {CRF} for real-time road-object segmentation from
  3{D} {LiDAR} point cloud. In: 2018 IEEE International Conference on Robotics
  and Automation (ICRA), pp 1887--1893

\bibitem[{Wu et~al.(2019)Wu, Zhou, Zhao, Yue, and Keutzer}]{wu2019squeezesegv2}
Wu B, Zhou X, Zhao S, Yue X, Keutzer K (2019) Squeezesegv2: Improved model
  structure and unsupervised domain adaptation for road-object segmentation
  from a {LiDAR} point cloud. In: 2019 International Conference on Robotics and
  Automation (ICRA), pp 4376--4382

\bibitem[{Wu et~al.(2018{\natexlab{b}})Wu, Qi, and Fuxin}]{wu2018pointconv}
Wu W, Qi Z, Fuxin L (2018{\natexlab{b}}) {PointConv}: Deep convolutional
  networks on 3{D} point clouds. In: Proceedings of the IEEE/CVF International
  Conference on Computer Vision, pp 9621--9630

\bibitem[{Wu et~al.(2015)Wu, Song, Khosla, Yu, Zhang, Tang, and
  Xiao}]{3d_shapenets}
Wu Z, Song S, Khosla A, Yu F, Zhang L, Tang X, Xiao J (2015) 3{D} {ShapeNets}:
  A deep representation for volumetric shapes. In: Proceedings of the IEEE/CVF
  International Conference on Computer Vision, pp 1912--1920

\bibitem[{Xie et~al.(2020)Xie, Gu, Guo, Qi, Guibas, and
  Litany}]{xie2020pointcontrast}
Xie S, Gu J, Guo D, Qi CR, Guibas L, Litany O (2020) Pointcontrast:
  Unsupervised pre-training for 3{D} point cloud understanding. In: European
  Conference on Computer Vision, pp 574--591

\bibitem[{Xu et~al.(2020)Xu, Wu, Wang, Zhan, Vajda, Keutzer, and
  Tomizuka}]{xu2020squeezesegv3}
Xu C, Wu B, Wang Z, Zhan W, Vajda P, Keutzer K, Tomizuka M (2020)
  Squeeze{S}eg{V}3: Spatially-adaptive convolution for efficient point-cloud
  segmentation. In: Proceedings of the European Conference on Computer Vision
  (ECCV), pp 1--19

\bibitem[{Xu and Lee(2020)}]{xu2020weakly}
Xu X, Lee GH (2020) Weakly supervised semantic point cloud segmentation:
  Towards 10x fewer labels. In: Proceedings of the IEEE/CVF Conference on
  Computer Vision and Pattern Recognition, pp 13706--13715

\bibitem[{Yan et~al.(2020)Yan, Zheng, Li, Wang, and Cui}]{yan2020pointasnl}
Yan X, Zheng C, Li Z, Wang S, Cui S (2020) {PointASNL}: Robust point clouds
  processing using nonlocal neural networks with adaptive sampling. In:
  Proceedings of the IEEE/CVF International Conference on Computer Vision, pp
  5589--5598

\bibitem[{Yang et~al.(2019)Yang, Wang, Clark, Hu, Wang, Markham, and
  Trigoni}]{3dbonet}
Yang B, Wang J, Clark R, Hu Q, Wang S, Markham A, Trigoni N (2019) Learning
  object bounding boxes for 3{D} instance segmentation on point clouds.
  Advances in Neural Information Processing Systems

\bibitem[{Ye et~al.(2018)Ye, Li, Huang, Du, and Zhang}]{3PRNN}
Ye X, Li J, Huang H, Du L, Zhang X (2018) {3D} recurrent neural networks with
  context fusion for point cloud semantic segmentation. In: Proceedings of the
  European Conference on Computer Vision (ECCV)

\bibitem[{Ye et~al.(2020)Ye, Xu, Huang, Tong, Li, Liu, Luan, Hoegner, and
  Stilla}]{ye2020lasdu}
Ye Z, Xu Y, Huang R, Tong X, Li X, Liu X, Luan K, Hoegner L, Stilla U (2020)
  {LASDU}: A large-scale aerial {LiDAR} dataset for semantic labeling in dense
  urban areas. ISPRS International Journal of Geo-Information 9(7):450

\bibitem[{Yi et~al.(2016)Yi, Kim, Ceylan, Shen, Yan, Su, Lu, Huang, Sheffer,
  and Guibas}]{ShapePartNet}
Yi L, Kim VG, Ceylan D, Shen IC, Yan M, Su H, Lu C, Huang Q, Sheffer A, Guibas
  L (2016) A scalable active framework for region annotation in 3{D} shape
  collections. ACM Transactions on Graphics (TOG) 35(6):1--12

\bibitem[{Yuan et~al.(2018)Yuan, Khot, Held, Mertz, and Hebert}]{yuan2018pcn}
Yuan W, Khot T, Held D, Mertz C, Hebert M (2018) {PCN}: Point completion
  network. In: 2018 International Conference on 3D Vision (3DV), pp 728--737

\bibitem[{Zhang et~al.(2020)Zhang, Zhou, David, Yue, Xi, Gong, and
  Foroosh}]{zhang2020polarnet}
Zhang Y, Zhou Z, David P, Yue X, Xi Z, Gong B, Foroosh H (2020) {PolarNet}: An
  improved grid representation for online {LiDAR} point clouds semantic
  segmentation. In: Proceedings of the IEEE/CVF International Conference on
  Computer Vision, pp 9601--9610

\bibitem[{Zhang et~al.(2018)Zhang, Gerke, Vosselman, and Yang}]{zhang2018patch}
Zhang Z, Gerke M, Vosselman G, Yang MY (2018) A patch-based method for the
  evaluation of dense image matching quality. International journal of applied
  earth observation and geoinformation 70:25--34

\bibitem[{Zhang et~al.(2019)Zhang, Hua, and Yeung}]{zhang2019shellnet}
Zhang Z, Hua BS, Yeung SK (2019) Shell{N}et: Efficient point cloud
  convolutional neural networks using concentric shells statistics. Proceedings
  of the IEEE/CVF International Conference on Computer Vision pp 1607--1616

\bibitem[{Zhang et~al.(2021)Zhang, Girdhar, Joulin, and Misra}]{zhang2021self}
Zhang Z, Girdhar R, Joulin A, Misra I (2021) Self-supervised pretraining of
  3{D} features on any point-cloud. arXiv preprint arXiv:210102691

\bibitem[{Zhao et~al.(2020)Zhao, Jiang, Jia, Torr, and
  Koltun}]{Point_transformer}
Zhao H, Jiang L, Jia J, Torr P, Koltun V (2020) Point transformer. arXiv
  preprint arXiv:201209164

\bibitem[{Zhou et~al.(2017)Zhou, Zhao, Puig, Fidler, Barriuso, and
  Torralba}]{scenenn}
Zhou B, Zhao H, Puig X, Fidler S, Barriuso A, Torralba A (2017) Scene parsing
  through {ADE20K} dataset. In: Proceedings of the IEEE/CVF International
  Conference on Computer Vision, pp 633--641

\bibitem[{Zhou and Tuzel(2018)}]{zhou2018voxelnet}
Zhou Y, Tuzel O (2018) {VoxelNet}: End-to-end learning for point cloud based
  3{D} object detection. In: Proceedings of the IEEE/CVF International
  Conference on Computer Vision, pp 4490--4499

\bibitem[{Zhu et~al.(2021)Zhu, Zhou, Wang, Hong, Ma, Li, Li, and
  Lin}]{cylinder3d}
Zhu X, Zhou H, Wang T, Hong F, Ma Y, Li W, Li H, Lin D (2021) Cylindrical and
  asymmetrical 3{D} convolution networks for {LiDAR} segmentation. In:
  Proceedings of the IEEE/CVF Conference on Computer Vision and Pattern
  Recognition

\bibitem[{Zolanvari et~al.(2019)Zolanvari, Ruano, Rana, Cummins, da~Silva,
  Rahbar, and Smolic}]{zolanvari2019dublincity}
Zolanvari S, Ruano S, Rana A, Cummins A, da~Silva RE, Rahbar M, Smolic A (2019)
  Dublin{C}ity: Annotated {LiDAR} point cloud and its applications. In: British
  Machine Vision Conference

\end{thebibliography}

\end{document}